\def\OMMTArxivVersion{1}
\documentclass{llncs}

\usepackage[T1]{fontenc}
\usepackage{graphicx}
\usepackage{amsmath,amssymb,mathtools}
\usepackage{microtype}
\usepackage{xspace}
\usepackage{bm}
\usepackage{algorithm}
\usepackage{algpseudocode}
\usepackage{hyperref}
\usepackage{cleveref}
\usepackage{bbding}

\newif\ifwithappendix
\ifdefined\OMMTArxivVersion
  \withappendixtrue
\else
  \withappendixfalse
\fi
\newcommand{\FullVersionReference}{the full version on arXiv}
\newcommand{\ProofLocation}[1]{%
  \ifwithappendix\Cref{#1}%
  \else\FullVersionReference\fi}

\newcommand{\E}{\mathbb{E}}
\newcommand{\Prob}{\mathbb{P}}
\newcommand{\Sn}{\mathfrak{S}_n}
\newcommand{\cS}{\mathcal{S}}

\newcommand{\cD}{\mathcal{D}}

\newcommand{\bell}{\boldsymbol{\ell}}

\newcommand{\ALG}{\mathrm{ALG}}
\newcommand{\OPT}{\mathrm{OPT}}
\newcommand{\Reg}{\mathrm{Reg}}

\newcommand{\OMMT}{\textsc{OMMT}\xspace}

\newcommand{\FTL}{\textsc{FTL}\xspace}

\spnewtheorem{assumption}{Assumption}{\bfseries}{\itshape}
\spnewtheorem{myclaim}{Claim}{\bfseries}{\itshape}

\crefname{assumption}{Assumption}{Assumptions}
\Crefname{assumption}{Assumption}{Assumptions}
\crefname{myclaim}{Claim}{Claims}
\Crefname{myclaim}{Claim}{Claims}

\begin{document}

\title{Online Learning for Minimizing Maximum Tardiness in Single-Machine Scheduling}
\titlerunning{Online Learning for Minimizing Maximum Tardiness}

\author{
Yiyang Chen\inst{1,2}\orcidID{0009-0007-8609-6814}\Envelope
\and
Kohei Hatano\inst{1,2}\orcidID{0000-0002-1536-1269}
\and
Eiji Takimoto\inst{1}\orcidID{0000-0001-9542-2553}
}

\authorrunning{Y. Chen, K. Hatano, and E. Takimoto}

\institute{
Kyushu University, Fukuoka, Japan\\
\email{chen.yiyang.894@s.kyushu-u.ac.jp}\\
\email{\{hatano,eiji\}@inf.kyushu-u.ac.jp}
\and
RIKEN AIP, Japan
}

\maketitle              

\begin{abstract}
We study online learning for minimizing maximum tardiness (\OMMT)
in single-machine scheduling.
In each round, the learner selects a permutation of \(n\) fixed jobs,
representing a schedule.
The environment then reveals a processing time and a due date for
each job, and the learner incurs a vector-valued loss whose components
correspond to the tardiness of the individual jobs.
The goal of the learner is to minimize \(\alpha\)-regret,
the difference between the learner's maximum cumulative tardiness
and \(\alpha\) times that of the best fixed permutation in hindsight,
for as small \(\alpha\) as possible.
We investigate OMMT under both stochastic and adversarial environments.
In the stochastic setting,  
we prove that the expected \(1\)-regret of the
Follow-the-Leader (FTL) algorithm is
\(
  O\!\left(\frac{n\log n}{\Delta^2}\right)
\),
together with a corresponding high-probability bound,
where \(\Delta\) is a distribution-dependent gap parameter.
At round \(t \in \{1,\dots,T\}\), FTL runs in \(O(tn^2)\) time and
\(O(tn)\) space.
In the adversarial setting, we prove that, for any
\(0<c<\log_2 n/16\), the \(\alpha\)-regret is \(\Omega(T)\) whenever
\(\alpha<\log_2 n/16-c\).
We then reduce OMMT to a surrogate online learning problem
and apply Follow-the-Perturbed-Leader (FTPL), obtaining an expected
\(\alpha\)-regret bound of \(O(n^2\sqrt{T})\) for
\(\alpha=O(\log n)\).
Thus, the logarithmic dependence of \(\alpha\) on \(n\) is tight up to
constant factors.
Each FTPL update admits a mixed-integer programming formulation
whose size is polynomial in \(n\) and \(t\).

\end{abstract}

\section{Introduction}
\label{sec:introduction}

Scheduling is a fundamental class of optimization problems that arises in
manufacturing, computing systems, logistics, and many other operational
settings.
Offline scheduling problems have been extensively studied~\cite{Pinedo2016}.
Among these, the minimization of the maximum tardiness of jobs on a single
machine is a classical problem.
Given the processing times and due dates of jobs in \([n]\), the goal is to
find a schedule---a permutation of the \(n\) jobs---that minimizes the
maximum tardiness, defined as the positive delay of a job's completion time
beyond its due date.
This problem can be solved by Lawler's Lowest Cost Last (LCL)
algorithm~\cite{Lawler1973} in \(O(n^2)\) time.

In this work, we consider a repeated online learning formulation of
this classical problem, which we refer to as
Online learning for Minimizing Maximum Tardiness (OMMT).
The learner repeatedly schedules the same set of \(n\) jobs under
round-dependent processing times and due dates.
At each round \(t=1,\ldots,T\),
(i) the learner selects a permutation \(\sigma^{(t)}\),
(ii) the environment reveals the processing times and due dates of the
jobs, and
(iii) the learner incurs the resulting job-wise tardiness vector.
The learner is evaluated by \(\alpha\)-regret, which compares the
\(\ell_\infty\)-norm of its cumulative tardiness vector with
\(\alpha\) times that of the best fixed permutation in hindsight.
This model captures repeated scheduling settings in which the set of
jobs is known in advance, but their realized processing requirements
and due dates vary over time and are not available when the scheduling
decision is made.
For example, the same collection of tasks may need to be scheduled
repeatedly under changing workloads, machine conditions, or
time-dependent deadlines.
In such settings, a scheduler must learn from past observations while
remaining robust to variation in future instances.

A key challenge is that OMMT does not fit directly into standard online
linear or convex optimization frameworks.
In conventional combinatorial online learning, losses are often linear
in a vector representation of the learner's combinatorial decision, and
performance is evaluated through cumulative scalar losses.
This linear structure underlies the tractability of many combinatorial
online learning problems.
In contrast, OMMT has a vector-valued and nonlinear loss structure, and
performance is measured by the \(\ell_\infty\)-norm of the cumulative
loss vector rather than by a sum of scalar losses.
OMMT is also related to online learning problems with global cost
functions, such as online load balancing
~\cite{EvenDarKleinbergMannorMansour2009,EvenDarMannorMansour2010,LiuHatanoTakimoto2021},
and to the Online Learning with Vector Costs (OLVC) framework of
Kesselheim and Singla~\cite{KesselheimSingla2020}.
However, OMMT additionally involves scheduling constraints and a
decision space consisting of permutations.

For the stochastic environment, where the instances are drawn
independently and identically from an unknown distribution, we consider
the classical Follow-the-Leader (FTL) algorithm~\cite{CesaBianchiLugosi2006}
under a distribution-dependent gap condition.
When the population-optimal permutation is unique, we show that its
expected \(1\)-regret is
\(
  O\!\left(
    \frac{n\log n}{\Delta^2}
  \right),
\)
where \(\Delta\) is the gap between the optimal and second-best
population values.
We also establish a corresponding high-probability bound.
The analysis extends to the case of multiple population-optimal
permutations, for which we obtain
\(
  O\!\left(
    \sqrt{Tn\log n}
    +
    \frac{n\log n}{\Delta^2}
  \right)
\)
expected regret.
The permutation selected by FTL can be computed by an extension of
Lawler's LCL algorithm, which runs in \(O(tn^2)\) time and uses
\(O(tn)\) space at round \(t\).

For the adversarial setting, 
where feedback is chosen adversarially based on the learner's past decisions 
(i.e., the adversary is adaptive),
we first prove that, for any
\(0<c<\log_2 n/16\), the \(\alpha\)-regret is \(\Omega(T)\) whenever
\(\alpha<\log_2 n/16-c\).
Thus, sublinear \(\alpha\)-regret is impossible when the multiplicative
factor \(\alpha\) is below this logarithmic threshold.
We then obtain an expected \(\alpha\)-regret bound of
\(O(n^2\sqrt{T})\) for some \(\alpha=O(\log n)\).
Together, these results show that the logarithmic dependence of
\(\alpha\) on \(n\) is tight up to constant factors.
Our adversarial algorithm is based on a reduction to a scalar surrogate
online learning problem.
We apply the OLVC reduction of
Kesselheim and Singla~\cite{KesselheimSingla2020} to the nonlinear
vector losses arising in OMMT.
We then apply Follow-the-Perturbed-Leader
(FTPL)~\cite{SuggalaNetrapalli2020} to the surrogate problem.
Each FTPL update admits a mixed-integer programming (MIP) formulation with
\(O(n^2+nt)\) variables and \(O(n^3+nt)\) linear constraints.
Although this formulation has polynomial size, 
it is an open question whether 
the same regret guarantee can be achieved in polynomial time.
We also investigate whether the classical
Follow-the-Leader/Be-the-Leader (FTL--BTL) lemma
~\cite{CesaBianchiLugosi2006,Hazan2016,KalaiVempala2005}
admits an analogous extension to the global \(\ell_\infty\)-objective.
Such an extension, together with a suitable stability argument, would
provide a direct route to analyzing FTL-type algorithms.
However, we give a counterexample showing that a natural multiplicative
extension of the FTL--BTL lemma fails in general.
Beyond its relevance to OMMT, this negative result may also be of
independent interest for online learning with vector-valued losses and
global objectives.

Our results also have theoretical implications for online learning.
A central question is whether an algorithm for an offline optimization
problem can be used to construct an online learning algorithm, a process
known as offline-to-online conversion.
For bounded linear losses, FTPL provides such a conversion: with access
to an appropriate offline optimization oracle, it achieves sublinear
\(1\)-regret, that is, \(o(T)\) regret.
Our adversarial lower bound shows a limitation of this approach for
global objectives defined by nonlinear functions of cumulative loss
vectors. Even when the corresponding offline problem can be solved
exactly, sublinear \(\alpha\)-regret is impossible below the logarithmic
threshold specified in \Cref{thm:adversarial-lower-bound}.

\subsection{Related Work}
\label{sec:related-work}

Several online learning formulations of single-machine scheduling have
been studied from the regret-minimization viewpoint. 
Yasutake et al.~\cite{Yasutake2011} studied the
online problem of minimizing the total completion time and obtained a regret
bound of \(O(n^2\sqrt{T\log n})\).  
For the same objective, 
Suehiro et al.~\cite{Suehiro2012} and Ailon~\cite{Ailon2014}
obtained regret bounds of \(O(n^2\sqrt{T})\).  
Fujita et al.~\cite{Fujita2015} studied the precedence-constrained version of this problem 
and proved an \(\alpha\)-regret bound of
\(O(n^2\sqrt{T})\), with $\alpha=2-2/(n+2)$.
These works are closely related to Online Linear Optimization (OLO) over permutations.  
\OMMT is structurally different:
each round generates a job-wise tardiness vector, and the learner is evaluated
by a global loss function of the cumulative vector.

Online learning problems with vector-valued losses and global loss functions have
also been studied, most notably in Online Load Balancing
(OLB)~\cite{EvenDarKleinbergMannorMansour2009,EvenDarMannorMansour2010,LiuHatanoTakimoto2021}.  
Kesselheim and Singla~\cite{KesselheimSingla2020} further developed the OLVC framework.
These works are highly relevant to \OMMT, but their loss structures
are different from that of \OMMT.
OLB and OLVC typically consider linear vector losses, whereas in
\OMMT the vector loss induced by a permutation is nonlinear.

\section{Preliminaries}
\label{sec:preliminaries}

We first introduce the classical single-machine maximum tardiness
minimization problem that underlies \OMMT.
In scheduling theory, problems are commonly described by the three-field notation
$\alpha|\beta|\gamma$~\cite{Pinedo2016},  
where \(\alpha\), \(\beta\), and \(\gamma\) specify the machine
environment, processing constraints and job characteristics, and
objective function, respectively.
The single-machine maximum tardiness minimization problem is denoted by
\(1||T_{\max}\).

More specifically, consider \(n\) jobs indexed by
\([n]:=\{1,\ldots,n\}\).
Each job \(j\in[n]\) has a processing time
\(p_j\in\mathbb{R}_{\ge0}\) and a due date
\(d_j\in\mathbb{R}_{\ge0}\).
All jobs are available at time zero and are processed nonpreemptively
on a single machine.
Hence, a non-idling schedule can be represented by a permutation of
the jobs.
Let \(\Sn\) denote the set of all permutations of \([n]\).
For \(\sigma\in\Sn\), we use the position convention:
\(\sigma(j)\) denotes the position of job \(j\) in the processing order,
so a smaller \(\sigma(j)\) means earlier processing.
For a permutation \(\sigma\), the completion time and tardiness of job
\(j\) are
\[
  C_j(\sigma;\bm p)
  :=
  \sum_{i:\,\sigma(i)\le\sigma(j)} p_i,
  \qquad
  T_j(\sigma;\bm p,\bm d)
  :=
  \bigl(C_j(\sigma;\bm p)-d_j\bigr)_+ .
\]
The maximum tardiness is
\(T_{\max}(\sigma;\bm p,\bm d)
:=\max_{j\in[n]}T_j(\sigma;\bm p,\bm d)\),
and the goal is to find a permutation minimizing \(T_{\max}\).

The problem \(1||T_{\max}\) is a special case of
\(1|\mathrm{prec}|h_{\max}\), where each job \(j\) has a nondecreasing
cost function \(h_j\) of its completion time and the objective is to
minimize the maximum job cost.
Lawler's Lowest Cost Last (LCL) algorithm~\cite{Lawler1973}
solves \(1|\mathrm{prec}|h_{\max}\) by a backward greedy procedure in
\(O(n^2)\) time.
By taking \(h_j(x)=(x-d_j)_+\) and an empty precedence relation,
\(1||T_{\max}\) is recovered as a special case and can therefore also
be solved in \(O(n^2)\) time by LCL.

\subsection{The \OMMT Model}
\label{sec:ommt}

We now formalize \OMMT over a fixed set of \(n\) jobs.
At each round \(t=1,\ldots,T\), the learner chooses a permutation
\(\sigma^{(t)}\in\Sn\) before observing the processing-time vector
\(\bm p^{(t)}\in\mathbb{R}_{\ge0}^n\) and the due-date vector
\(\bm d^{(t)}\in\mathbb{R}_{\ge0}^n\).
After \((\bm p^{(t)},\bm d^{(t)})\) is revealed, the learner incurs
a vector-valued loss determined by the job tardinesses under
\(\sigma^{(t)}\).

For any \(\sigma\in\Sn\), the completion time and normalized loss of job
\(j\) at round \(t\) are
\[
  C_j^{(t)}(\sigma)
  :=
  \sum_{i:\,\sigma(i)\le\sigma(j)} p_i^{(t)},
  \qquad
  \ell_j^{(t)}(\sigma)
  :=
  \frac{1}{n}
  \bigl(C_j^{(t)}(\sigma)-d_j^{(t)}\bigr)_+ ,
\]
and let
\(
  \bell^{(t)}(\sigma)
  :=
  (\ell_1^{(t)}(\sigma),\ldots,\ell_n^{(t)}(\sigma))
\).
We assume the instances are scaled so that
\(\bell^{(t)}(\sigma)\in[0,1]^n\) for every
\(t\in[T]\) and \(\sigma\in\Sn\).

The learner is evaluated by the
\(\ell_\infty\)-norm of its cumulative loss vector. Define
\[
  \ALG_T
  :=
  \left\|
    \sum_{t=1}^{T}
    \bell^{(t)}(\sigma^{(t)})
  \right\|_\infty,
  \qquad
  \OPT_T
  :=
  \min_{\sigma\in\Sn}
  \left\|
    \sum_{t=1}^{T}
    \bell^{(t)}(\sigma)
  \right\|_\infty .
\]
Here, \(\ALG_T\) is the learner's cumulative cost, while
\(\OPT_T\) is the best cumulative cost achievable in hindsight by a
single fixed permutation.

For \(\alpha\ge1\), we measure performance by the
\(\alpha\)-regret
\[
  \alpha\text{-}\Reg_T
  :=
  \ALG_T-\alpha\OPT_T.
\]
For \(\alpha=1\), we simply write
\(\Reg_T\).

\section{Stochastic \OMMT}
\label{sec:stochastic}

In the stochastic environment, the vectors
\(
  (\bm p^{(t)},\bm d^{(t)})
\),
\(t=1,\ldots,T\), are drawn independently and identically from an
unknown distribution \(\mathcal D\).
Consequently, for every fixed \(\sigma\in\Sn\), 
the corresponding vector losses
\(
  \bell^{(1)}(\sigma),\ldots,\bell^{(T)}(\sigma)
\)
are also independent and identically distributed
(i.i.d.).

\subsection{FTL}
\label{sec:stoch-ftl}

We apply the Follow-the-Leader (FTL) algorithm to stochastic \OMMT.
At \(t=1\), FTL selects an arbitrary permutation.
For every \(t\ge2\), based on the instances observed before round \(t\),
it selects
\begin{equation}
\label{eq:stoch-ftl}
  \sigma^{(t)}
  \in
  \operatorname*{arg\,min}_{\sigma\in\Sn}
  \left\|
    \sum_{s=1}^{t-1}\bell^{(s)}(\sigma)
  \right\|_\infty .
\end{equation}

To compute the permutation \(\sigma^{(t)}\) in \eqref{eq:stoch-ftl},
we use a backward greedy procedure, which we call
\emph{Extended LCL}; see \Cref{alg:extended-lcl}.
The procedure constructs the permutation backward.
At each step, let \(S\subseteq[n]\) denote the set of jobs whose
positions have not yet been assigned.
For each candidate \(i\in S\) and each observed instance
\(s\in\{1,\ldots,t-1\}\), if \(i\) is assigned the last remaining
position, then its completion time is
\(
  \sum_{j\in S}p_j^{(s)}
\).
Hence, its tardiness in that instance is
\(
  \bigl(\sum_{j\in S}p_j^{(s)}-d_i^{(s)}\bigr)_+
\).
Extended LCL assigns the last remaining position to a job minimizing
the sum of this quantity over all observed instances and then repeats
the same procedure for the remaining jobs.
The factor \(1/n\) in the \OMMT loss is common to all candidates and
thus does not affect the minimizer.

\begin{algorithm}[t]
\caption{Extended LCL at round \(t\)}
\label{alg:extended-lcl}
\begin{algorithmic}[1]
\Require Observed instances
\(
  \{(\bm p^{(s)},\bm d^{(s)})\}_{s=1}^{t-1}
\)
\State \(S\gets[n]\)
\While{\(S\neq\emptyset\)}
  \State
  \(
    i^\star
    \in
    \operatorname*{arg\,min}_{i\in S}
    \sum_{s=1}^{t-1}
    \left(
      \sum_{j\in S}p_j^{(s)}
      -
      d_i^{(s)}
    \right)_+
  \)
  \State \(\sigma^{(t)}(i^\star)\gets |S|\)
  \State \(S\gets S\setminus\{i^\star\}\)
\EndWhile
\State \Return \(\sigma^{(t)}\)
\end{algorithmic}
\end{algorithm}

When only one instance is considered, Extended LCL reduces to the
classical LCL algorithm.
Thus, it can be viewed as its cumulative extension.

\begin{proposition}
\label{prop:cumulative-lcl}
For every round \(t\ge2\), Extended LCL computes an exact minimizer
for the FTL update in \eqref{eq:stoch-ftl}.
It can be implemented in \(O(tn^2)\) time using \(O(tn)\) space.
\end{proposition}

The proof is given in
\ProofLocation{app:proof-cumulative-lcl}.

We next analyze the regret of FTL.
For each \(\sigma\in\Sn\), let
\(
  \bm\mu(\sigma):=\E_{\mathcal D}[\bell(\sigma)]
\).
Throughout the following analysis, we assume that the minimizer of
\(
  \|\bm\mu(\sigma)\|_\infty
\)
over \(\sigma\in\Sn\) is unique, and denote it by
\(\sigma^\star\).
Define the gap
\begin{equation}
\label{eq:unique-population-gap}
  \Delta
  :=
  \min_{\sigma\neq\sigma^\star}
  \left(
    \|\bm\mu(\sigma)\|_\infty
    -
    \|\bm\mu(\sigma^\star)\|_\infty
  \right).
\end{equation}
Since \(\Sn\) is finite, uniqueness of \(\sigma^\star\) implies
\(\Delta>0\).

We compare the permutations selected by FTL directly with the fixed
permutation \(\sigma^\star\).
The regret can then be decomposed as
\begin{equation}
\label{eq:unique-ftl-decomposition}
\begin{aligned}
  \Reg_T^{\FTL}
  &=
  \left\|
    \sum_{t=1}^{T}
    \bell^{(t)}(\sigma^{(t)})
  \right\|_\infty
  -
  \min_{\sigma\in\Sn}
  \left\|
    \sum_{t=1}^{T}
    \bell^{(t)}(\sigma)
  \right\|_\infty
  \\
  &\le
  \left\|
    \sum_{t=1}^{T}
    \bell^{(t)}(\sigma^\star)
  \right\|_\infty
  +
  \sum_{t=1}^{T}
  \left\|
    \bell^{(t)}(\sigma^{(t)})
    -
    \bell^{(t)}(\sigma^\star)
  \right\|_\infty
  -
  \OPT_T
  && \text{(a)}
  \\
  &\le
  \left\|
    \sum_{t=1}^{T}
    \bell^{(t)}(\sigma^\star)
  \right\|_\infty
  +
  \underbrace{
    \sum_{t=1}^{T}
    \mathbf 1\{\sigma^{(t)}\neq\sigma^\star\}
  }_{=:N_T}
  -
  \OPT_T
  && \text{(b)}
  \\
  &=
  N_T
  +
  \underbrace{
    \left\|
      \sum_{t=1}^{T}
      \bell^{(t)}(\sigma^\star)
    \right\|_\infty
    -
    \OPT_T
  }_{=:B_T}.
\end{aligned}
\end{equation}
For (a), we apply the triangle inequality.
For (b), since all loss coordinates lie in \([0,1]\),
\(
  \|\bell^{(t)}(\sigma^{(t)})-\bell^{(t)}(\sigma^\star)\|_\infty
  \le
  \mathbf 1\{\sigma^{(t)}\neq\sigma^\star\}
\)
for every \(t\).
It remains to control \(N_T\) and \(B_T\).
The following lemmas give the required bounds.

\begin{lemma}
\label{lem:unique-ftl-stabilization}
For every \(t\ge2\),
\(
  \Prob(\sigma^{(t)}\neq\sigma^\star)
  \le
  2nn!\exp(-(t-1)\Delta^2/8)
\),
and
\(
  \E[N_T]=O(n\log n/\Delta^2)
\).
Moreover, for every \(\delta\in(0,1)\), with probability at least
\(1-\delta\),
\(
  N_T
  =
  O((n\log n+\log(1/\delta)+\log(1/\Delta))/\Delta^2)
\).
\end{lemma}

The proof is given in
\ProofLocation{app:proof-unique-ftl-stabilization}.

\begin{lemma}
\label{lem:unique-benchmark}
\(
  \Prob(B_T>0)
  \le
  2nn!\exp(-T\Delta^2/8)
\),
and
\(
  \E[B_T]=O(n\log n/\Delta^2)
\).
Moreover, for every \(\delta\in(0,1)\), with probability at least
\(1-\delta\),
\(
  B_T
  =
  O((n\log n+\log(1/\delta))/\Delta^2)
\).
\end{lemma}

The proof is given in
\ProofLocation{app:proof-unique-benchmark}.

Combining the preceding lemmas with
\eqref{eq:unique-ftl-decomposition} gives the following expected and
high-probability regret guarantees.

\begin{theorem}
\label{thm:stoch-ftl-expectation}
For stochastic \OMMT,
the expected regret of FTL satisfies
\[
  \E[\Reg_T^{\FTL}]
  =
  O\!\left(
    \frac{n\log n}{\Delta^2}
  \right).
\]
\end{theorem}

\begin{theorem}
\label{thm:stoch-ftl-highprob}
For every \(\delta\in(0,1)\), 
with probability at least
\(1-\delta\),
the regret of FTL satisfies
\[
  \Reg_T^{\FTL}
  =
  O\!\left(
    \frac{
      n\log n+\log(1/\delta)+\log(1/\Delta)
    }{\Delta^2}
  \right).
\]
\end{theorem}

The analysis can also be extended when the minimizer of
\(
  \|\bm\mu(\sigma)\|_\infty
\)
is not unique.
Let
\(
  \cS^\star
  :=
  \operatorname*{arg\,min}_{\sigma\in\Sn}
  \|\bm\mu(\sigma)\|_\infty
\).
In this setting, \(\Delta\) denotes the gap between
\(
  \|\bm\mu(\sigma)\|_\infty
\)
for permutations \(\sigma\notin\cS^\star\) and the common minimum
value attained by permutations in \(\cS^\star\).
FTL satisfies
\(
  \E[\Reg_T^{\FTL}]
  =
  O(\sqrt{Tn\log n}+n\log n/\Delta^2)
\).

The additional \(\sqrt{T}\)-order term arises from our analysis of the
non-unique setting.
The difference from the unique case is that FTL may continue to switch
among different permutations in \(\cS^\star\), even after it no longer
selects permutations outside \(\cS^\star\).
Therefore, its decisions cannot be compared directly with a single
fixed permutation as in \eqref{eq:unique-ftl-decomposition}.
The analysis instead uses a round-dependent comparison sequence and
controls the associated stochastic deviations by concentration
arguments.
The resulting concentration terms introduce the
\(\sqrt{T}\)-order dependence.
Whether this dependence on \(T\) is necessary when the minimizer is
not unique remains open.
The complete analysis is given in
\ProofLocation{app:stochastic-multiple-optima}.

\section{Adversarial \OMMT}
\label{sec:adversarial}

We consider an adaptive adversarial environment. At round \(t\), the instance
\((\bm{p}^{(t)},\bm{d}^{(t)})\) may depend on the learner's past
decisions \(\sigma^{(1)},\ldots,\sigma^{(t-1)}\).

\subsection{Lower bound on \texorpdfstring{$\alpha$}{alpha}-regret}
\label{sec:adversarial-lower-bound}

The following result shows that sublinear \(\alpha\)-regret is possible
only if \(\alpha=\Omega(\log n)\).

\begin{theorem}
\label{thm:adversarial-lower-bound}
Let \(n=2^K\) and \(T\) be a multiple of 
\(K\in\mathbb{N}_{+}\).
For any 
\(
  0<c<\log_2 n/16
\),
if
\(
  \alpha
  <
  \log_2 n/16-c,
\)
then there exists an adversarial sequence such that
\[
  \alpha\text{-}\Reg_T
  \ge
  \frac{c}{2\log_2 n}\,T .
\]
\end{theorem}

\begin{proof}
Let
\(
  K:=\log_2 n
\)
and
\(
  L:=T/K
\).
We divide the horizon \(T\) into \(K\) phases, 
where phase \(k\) consists of
the \(L\) rounds
\(
  \mathcal T_k
  :=
  \{(k-1)L+1,\ldots,kL\}.
\)
The adversary maintains nested active sets
\[
  [n]=A_1\supset A_2\supset\cdots\supset A_{K+1},
  \qquad
  |A_k|=:m_k=\frac{n}{2^{k-1}}.
\]

At each round \(t\in\mathcal T_k\), 
the adversary chooses a subset
\(B_t\subseteq A_k\) with \(|B_t|=m_k/2\), and set
\[
  p_i^{(t)}
  =
  \begin{cases}
    n/m_k, & i\in B_t,\\
    0,     & i\notin B_t,
  \end{cases}
  \qquad
  d_i^{(t)}
  =
  \begin{cases}
    n/4, & i\in B_t,\\
    n,   & i\notin B_t.
  \end{cases}
\]
For the jobs in \(B_t\), the losses are
\(\bigl(j/m_k-1/4\bigr)_+\), \(j=1,\ldots,m_k/2\), up to reordering. 
Therefore,
\begin{equation}
\label{eq:phase-loss-lower}
  \sum_{i\in A_k}
  \ell_i^{(t)}(\sigma^{(t)})
  =
  \sum_{j=1}^{m_k/2}
  \left(
    \frac{j}{m_k}-\frac14
  \right)_+
  \ge
  \frac{m_k}{32}.
\end{equation}

For each \(i\in[n]\), let
\(
  U_{i,k}:=\sum_{s=1}^{kL}\ell_i^{(s)}(\sigma^{(s)})
\)
and
\(
  U_k:=m_k^{-1}\sum_{i\in A_k}U_{i,k}
\).
After phase \(k\), the adversary chooses \(A_{k+1}\) as the
\(m_k/2\) jobs in \(A_k\) with the largest cumulative losses
\(U_{i,k}\).

Since \(A_k\) is the larger-loss half of \(A_{k-1}\), the average
cumulative loss over \(A_k\) before phase \(k\) is at least
\(U_{k-1}\).
Moreover, summing \eqref{eq:phase-loss-lower} over the \(L\) rounds of
phase \(k\) adds at least \(Lm_k/32\) total loss to \(A_k\).
Hence
\(
  U_k
  \ge
  U_{k-1}+L/32.
\)
Iterating over all \(K\) phases gives
\(
  U_K\ge KL/32
\).
At the end of phase \(K\), \(A_{K+1}\) is the larger-loss half of
\(A_K\) and contains a single job. Its cumulative loss is therefore at
least \(U_K\), and thus
\begin{equation}
\label{eq:alg-adversarial-lower}
  \ALG_T
  \ge
  \frac{KL}{32}
  =
  \frac{T}{32}.
\end{equation}

We next upper bound \(\OPT_T\).
For \(h=1,\ldots,K\), define
\(D_h:=A_h\setminus A_{h+1}\),
and
\(D_{K+1}:=A_{K+1}\).
Consider the fixed permutation
\(
  \sigma^\star
  =
  (D_{K+1},D_K,\ldots,D_1),
\)
with an arbitrary ordering within each block.

Fix a job \(i\in D_h\).
For every phase \(k>h\), we have \(i\notin A_k\), so
\(p_i^{(t)}=0\) and \(i\) incurs no loss.
Now consider a phase \(k\le h-2\).
If \(i\notin B_t\), then \(d_i^{(t)}=n\), while the total processing
time in the round is \(n/2\), so again its loss is zero.
If \(i\in B_t\), all jobs scheduled no later than \(i\) under
\(\sigma^\star\) that have positive processing time belong to
\(
  D_{K+1}\cup D_K\cup\cdots\cup D_h
  =
  A_h.
\)
There are at most \(m_h\) such jobs. Therefore,
\(
  C_i(\sigma^\star;\bm p^{(t)})
  \le
  n m_h/m_k
  \le
  n/4
  =
  d_i^{(t)}
\),
where the second inequality follows from
\(
  m_h/m_k=2^{k-h}\le1/4
\).
Thus \(i\) also incurs zero loss in every phase \(k\le h-2\).

Consequently, a job in \(D_h\) can incur positive loss only in the
phases
\(
  \{h-1,h\}\cap[K]
\).
In any round its loss is at most \(1/4\), because the total processing
time of the jobs in \(B_t\) is \(n/2\) and their due date is \(n/4\).
Hence every job has cumulative loss at most \(L/2\) under
\(\sigma^\star\). It follows that
\begin{equation}
\label{eq:opt-adversarial-upper}
  \OPT_T
  \le
  \left\|
    \sum_{t=1}^{T}
    \bell^{(t)}(\sigma^\star)
  \right\|_\infty
  \le
  \frac{L}{2}
  =
  \frac{T}{2\log_2 n}.
\end{equation}

Combining
\eqref{eq:alg-adversarial-lower} and
\eqref{eq:opt-adversarial-upper}, we obtain
\[
  \ALG_T-\alpha\OPT_T
  \ge
  \frac{T}{32}
  -
  \alpha\frac{T}{2\log_2 n}
  \ge
  \frac{c}{2\log_2 n}\,T,
\]
where the last inequality uses
\(
  \alpha<\log_2 n/16-c
\).
Thus, sublinear \(\alpha\)-regret requires
\(
  \alpha\ge\log_2 n/16-o(1)
\),
which in particular implies
\(
  \alpha=\Omega(\log n)
\).
\end{proof}

\subsection{Failure of an Extended FTL--BTL Lemma}
\label{sec:failure-ftl-btl}

We first recall the classical FTL--BTL lemma for scalar losses.
Let \(\mathcal X\) be an action set, and let
\(
  f^{(t)}:\mathcal X\to\mathbb R
\)
denote the scalar loss function at round \(t\).
For each \(t\ge1\), let
\(
  x^{(t+1)}
  \in
  \operatorname*{arg\,min}_{x\in\mathcal X}
  \sum_{s=1}^{t} f^{(s)}(x)
\)
be the leader used by FTL in round \(t+1\).

\begin{lemma}[FTL--BTL lemma~\cite{Hazan2016}]
\label{lem:scalar-ftl-btl}
For every \(u\in\mathcal X\),
\[
  \sum_{t=1}^{T} f^{(t)}(x^{(t+1)})
  \le
  \sum_{t=1}^{T} f^{(t)}(u).
\]
\end{lemma}

This lemma is a standard ingredient in the analysis of FTL.
We ask whether an analogous property can hold for the cumulative
vector-valued objective in \OMMT.

Let
\(
  x_\infty^{(t+1)}
  \in
  \operatorname*{arg\,min}_{x\in\mathcal X}
  \left\|
    \sum_{s=1}^{t}\bell^{(s)}(x)
  \right\|_\infty .
\)
We refer to \(x_\infty^{(t+1)}\) as the
\(\ell_\infty\)-leader used by FTL in round \(t+1\).

A natural extension of \Cref{lem:scalar-ftl-btl} would allow a
multiplicative factor \(\alpha\ge1\) and require
\begin{equation}
\label{eq:extended-ftl-btl}
  \left\|
    \sum_{t=1}^{T}
    \bell^{(t)}(x_\infty^{(t+1)})
  \right\|_\infty
  \le
  \alpha
  \left\|
    \sum_{t=1}^{T}\bell^{(t)}(u)
  \right\|_\infty
\end{equation}
for every fixed \(u\in\mathcal X\).
The relevant question is whether such an inequality can hold with a
small factor \(\alpha\) to yield a meaningful
\(\alpha\)-regret guarantee.
OMMT is a special case of this setting, by taking
\(\mathcal X=\Sn\) and
\(\bell^{(t)}(\sigma)\) to be the tardiness-loss vector.
The following result shows that no useful 
extension of the form \eqref{eq:extended-ftl-btl} is possible in
general.

\begin{proposition}
\label{prop:max-btl-counterexample}
There exists a two-dimensional nonnegative vector-loss instance with
unique \(\ell_\infty\)-leaders for which
\eqref{eq:extended-ftl-btl} requires \(\alpha=\Omega(T)\).
\end{proposition}

The proof is given in
\ProofLocation{app:proof-max-btl-counterexample}.

We also consider the corresponding question for other
\(\ell_p\)-norm objectives: for finite \(p\in(1,\infty)\), the required
factor can grow as \(\Omega(T^{1-1/p})\), whereas for \(p=1\) the
classical scalar FTL--BTL lemma applies.
Further details and some other related observations are given in
\ProofLocation{app:additional-observations}.

\subsection{Reduction to a Surrogate Online Learning Problem}
\label{sec:olvc-reduction}

We use the reduction developed by
Kesselheim and Singla~\cite{KesselheimSingla2020} for
Online Learning with Vector Costs (OLVC).
In the original OLVC formulation, there are \(m\) actions.
At each round \(t\), the learner chooses a distribution
\(
  \bm{x}^{(t)}\in\Delta_m
\).
The environment then reveals a cost matrix
\(
  C^{(t)}\in[0,1]^{d\times m}
\),
and the learner incurs the vector cost
\(
  C^{(t)}\bm{x}^{(t)}
\).
The learner's cumulative vector cost is
\(
  \sum_{t=1}^{T} C^{(t)}\bm{x}^{(t)},
\)
whose performance in OLVC is measured by its
\(\ell_p\)-norm for \(p\in[1,\infty]\).
Here we only consider \(p=\infty\), so the objective becomes
\(
  \left\|
    \sum_{t=1}^{T} C^{(t)}\bm{x}^{(t)}
  \right\|_\infty,
\)
compared with the best fixed distribution in hindsight.

Although OLVC is originally formulated with linear vector costs, the
reduction itself only requires that each action induces a
nonnegative vector loss in \([0,1]^d\).
This includes the nonlinear vector losses arising in \OMMT.
We use the OLVC reduction in this more general form.
Let \(\mathcal X\) be the action set, and let
\(
  \bm{v}^{(t)}:\mathcal X\to[0,1]^d
\)
denote the vector-loss function at round \(t\).
For a sequence of decisions
\(x^{(1)},\ldots,x^{(T)}\), define
\[
  \ALG_T^{\mathrm{OLVC}}
  :=
  \left\|
    \sum_{t=1}^{T}
    \bm{v}^{(t)}(x^{(t)})
  \right\|_\infty ,
  \qquad
  \OPT_T^{\mathrm{OLVC}}
  :=
  \min_{x\in\mathcal X}
  \left\|
    \sum_{t=1}^{T}
    \bm{v}^{(t)}(x)
  \right\|_\infty .
\]

We next describe the reduction to a scalar online learning problem.
For \(\eta>0\), define the smooth-max function
\(
  \psi_\eta(\bm{u})
  :=
  \frac{1}{\eta}
  \log
  \left(
    \sum_{i=1}^{d} e^{\eta u_i}
  \right),
  \bm{u}\in\mathbb R_{\ge0}^{d}.
\)
Let
\(
  \bm V^{(t)}
  :=
  \sum_{s=1}^{t}\bm v^{(s)}(x^{(s)})
\) 
and
\(
  \bm V^{(0)}:=\bm 0.
\)
At the beginning of round \(t\), define
\(
  \bm{g}^{(t)}
  :=
  \nabla\psi_\eta(\bm{V}^{(t-1)})
  \in\Delta_d.
\)
After the vector-loss function \(\bm{v}^{(t)}\) is revealed, define
the scalar surrogate loss
\(
  f^{(t)}(x)
  :=
  \left\langle
    \bm{v}^{(t)}(x),
    \bm{g}^{(t)}
  \right\rangle .
\)
Thus, the reduction induces a scalar online learning problem over the
same action set \(\mathcal X\).

Suppose that the algorithm for the surrogate problem satisfies
\begin{equation}
\label{eq:surrogate-regret}
  \E\!\left[
    \sum_{t=1}^{T} f^{(t)}(x^{(t)})
    -
    \min_{x\in\mathcal X}
    \sum_{t=1}^{T} f^{(t)}(x)
  \right]
  \le
  R_T .
\end{equation}
Here the expectation is taken with respect to the randomness of the
online algorithm.
The detailed analysis of the reduction is given in
\ProofLocation{app:olvc-reduction}.
The reduction then yields the following guarantee.

\begin{theorem}
\label{thm:olvc-reduction}
Consider a fixed sequence of vector-loss functions
\(
  \bm{v}^{(t)}:\mathcal X\to[0,1]^d
\).
Fix constants \(0<\gamma<c<1\), and set
\[
  \eta
  :=
  \begin{cases}
    \displaystyle
    \min
    \left\{
      \ln(c/\gamma),
      \frac{\ln(1+\gamma)}
           {\OPT_T^{\mathrm{OLVC}}}
    \right\},
    & \OPT_T^{\mathrm{OLVC}}>0,
    \\[2ex]
    \ln(c/\gamma),
    & \OPT_T^{\mathrm{OLVC}}=0.
  \end{cases}
\]
If the algorithm for the surrogate problem satisfies
\eqref{eq:surrogate-regret}, then
\[
  \E\!\left[
    \ALG_T^{\mathrm{OLVC}}
  \right]
  \le
  \frac{
    \ln d+c/(1-c)
  }{
    \ln(1+\gamma)
  }
  \OPT_T^{\mathrm{OLVC}}
  +
  \frac{
    \ln d+c/(1-c)
  }{
    \ln(c/\gamma)
  }
  +
  \frac{c}{\gamma(1-c)}
  R_T .
\]
\end{theorem}

The proof is given in
\ProofLocation{app:proof-olvc-reduction}.
The multiplicative factor is still \(O(\log d)\), as in
Kesselheim and Singla~\cite{KesselheimSingla2020}, while the
coefficients are made explicit in the bound above.

\OMMT is a special case of OLVC with
\(d=n\), \(\mathcal X=\Sn\), and
\(\bm{v}^{(t)}(\sigma)=\bell^{(t)}(\sigma)\).
Taking \(\gamma=1/3\) and \(c=2/3\) in
\Cref{thm:olvc-reduction} gives the following result.

\begin{corollary}
\label{cor:ommt-reduction}
For \OMMT,
\[
  \E[\ALG_T]
  \le
  \frac{\ln n+2}{\ln(4/3)}\OPT_T
  +
  \frac{\ln n+2}{\ln 2}
  +
  6R_T .
\]
\end{corollary}

This yields an expected \(\alpha\)-regret bound with
\(\alpha \approx 3.5\ln n\) for large \(n\), up to the additive term
\(6R_T\).

It remains to choose an algorithm for the surrogate problem and to bound
\(R_T\). Treating the \(n!\) permutations as experts and applying
Hedge~\cite{LittlestoneWarmuth1994} gives
\(
  R_T=O(\sqrt{Tn\log n})
\),
but explicitly maintaining all \(n!\) experts is computationally infeasible.
We therefore apply FTPL to the surrogate problem.

\subsection{FTPL}
\label{sec:cm-ftpl}
We first introduce the comparison-matrix 
representation of a permutation.
For \(\sigma\in\Sn\), define the comparison matrix
\(\Pi(\sigma)\in\{0,1\}^{n\times n}\) by
\[
  \Pi_{ij}(\sigma)
  :=
  \mathbf{1}\{\sigma(i)\le\sigma(j)\},
  \qquad i,j\in[n].
\]
Thus \(\Pi_{ij}(\sigma)=1\) if job \(i\) is scheduled no later than job
\(j\), and in particular \(\Pi_{ii}(\sigma)=1\).
Let
\(
  \mathcal{P}_n
  :=
  \{\Pi(\sigma):\sigma\in\Sn\}
\).
Every \(\Pi\in\mathcal{P}_n\) satisfies
\[
  \Pi_{ii}=1,\qquad
  \Pi_{ij}+\Pi_{ji}=1\ (i\neq j),\qquad
  \Pi_{ij}+\Pi_{jk}-\Pi_{ik}\le1
  \quad(i,j,k\ \text{distinct}),
\]
together with \(\Pi_{ij}\in\{0,1\}\).

For \(\Pi=\Pi(\sigma)\), the completion time of job \(i\) is
\(
  C_i(\Pi;\bm{p})
  =
  \sum_{j=1}^{n}\Pi_{ji}p_j
\).
Hence, under the normalization used throughout the paper,
\(
  \ell_i(\Pi)
  =
  \frac{1}{n}
  \bigl(
    \sum_{j=1}^{n}\Pi_{ji}p_j-d_i
  \bigr)_+
\).
Given \(\bm g^{(t)}\in\Delta_n\), 
the round-\(t\) surrogate loss can be written as
\begin{equation}
\label{eq:cm-surrogate}
  f^{(t)}(\Pi)
  =
  \frac{1}{n}
  \sum_{i=1}^{n}
  g_i^{(t)}
  \left(
    \sum_{j=1}^{n}
    \Pi_{ji}p_j^{(t)}
    -
    d_i^{(t)}
  \right)_+ .
\end{equation}
For fixed \(\bm g^{(t)}\), \(f^{(t)}\) is convex in \(\Pi\).
However, since \(\mathcal P_n\) is discrete, the surrogate problem is
not a standard online convex optimization problem.

We apply FTPL to the surrogate problem.
At the beginning of round \(t\), independently draw a perturbation
matrix
\(
  \bm{\xi}^{(t)}=(\xi_{ij}^{(t)})_{i,j\in[n]}
\),
whose entries are independent and follow an exponential distribution
with rate parameter \(\rho>0\), i.e.,
\(
  \xi_{ij}^{(t)}\sim \mathrm{Exp}(\rho).
\)
FTPL then selects
\begin{equation}
\label{eq:cm-ftpl-update}
  \Pi^{(t)}
  \in
  \operatorname*{arg\,min}_{\Pi\in\mathcal{P}_n}
  \left\{
    \sum_{s=1}^{t-1}f^{(s)}(\Pi)
    -
    \left\langle
      \bm{\xi}^{(t)},\Pi
    \right\rangle
  \right\},
\end{equation}
where
\(
  \langle\bm{\xi}^{(t)},\Pi\rangle
  :=
  \sum_{i,j=1}^{n}\xi_{ij}^{(t)}\Pi_{ij}
\).

The FTPL update in \eqref{eq:cm-ftpl-update} admits the following MIP formulation:
\[
\begin{aligned}
\min_{\Pi,z}\quad
&
\frac{1}{n}
\sum_{s=1}^{t-1}
\sum_{i=1}^{n}
g_i^{(s)}z_i^{(s)}
-
\sum_{i,j=1}^{n}
\xi_{ij}^{(t)}\Pi_{ij}
\\
\text{s.t.}\quad
&
z_i^{(s)}
\ge
\sum_{j=1}^{n}
\Pi_{ji}p_j^{(s)}
-
d_i^{(s)},
\qquad
z_i^{(s)}\ge0,
\\
&
\Pi_{ii}=1,
\qquad
\Pi_{ij}+\Pi_{ji}=1
\quad(i\neq j),
\\
&
\Pi_{ij}+\Pi_{jk}-\Pi_{ik}\le1
\quad(i,j,k\text{ distinct}),
\qquad
\Pi_{ij}\in\{0,1\}.
\end{aligned}
\]
The variables \(z_i^{(s)}\) linearize the positive-part terms in
\eqref{eq:cm-surrogate}.
The formulation contains \(O(n^2+nt)\) variables and
\(O(n^3+nt)\) linear constraints.

We use the following FTPL regret bound.

\begin{theorem}[Suggala--Netrapalli~\cite{SuggalaNetrapalli2020}]
\label{thm:sn-ftpl}
Let
\(
  \mathcal X\subseteq\mathbb R^d
\)
have \(\ell_\infty\)-diameter \(D\), and suppose that each loss function
\(
  f^{(t)}:\mathcal X\to\mathbb R
\)
is \(L\)-Lipschitz with respect to the \(\ell_1\)-norm.
Consider FTPL with exact minimization in each update and independent
perturbations whose coordinates follow an exponential distribution
with rate \(\rho>0\).
Then the expected regret \(R_T\) in \eqref{eq:surrogate-regret} satisfies
\(
  R_T
  =
  O\!\left(
    \rho T d^2DL^2
    +
    dD/\rho
  \right).
\)
In particular, choosing
\(
  \rho
  =
  \frac{1}{L\sqrt{dT}}
\)
up to a constant factor gives
\[
  R_T
  =
  O\!\left(
    d^{3/2}DL\sqrt{T}
  \right).
\]

\end{theorem}

\begin{theorem}
\label{thm:cm-ftpl-regret}
For the surrogate problem of \OMMT, FTPL satisfies
\[
  R_T^{\mathrm{FTPL}}
  =
  O(n^2\sqrt{T}).
\]
\end{theorem}

\begin{proof}
The comparison-matrix domain lies in \(\mathbb{R}^{n^2}\), so
\(d=n^2\); its binary entries imply \(D\le1\).
As \(\bm{g}^{(t)}\in\Delta_n\) and \(x\mapsto x_+\) is
\(1\)-Lipschitz, the surrogate loss in \eqref{eq:cm-surrogate}
is \(1/n\)-Lipschitz with respect to the \(\ell_1\)-norm.
The claim follows from \Cref{thm:sn-ftpl}.
\end{proof}

Combining \Cref{cor:ommt-reduction} and
\Cref{thm:cm-ftpl-regret}, FTPL achieves
\(
  \E\!\left[
    \alpha\text{-}\Reg_T
  \right]
  =
  O(n^2\sqrt{T})
\)
for some
\(
  \alpha=O(\log n)
\).
Together with \Cref{thm:adversarial-lower-bound}, this shows that the
logarithmic dependence of \(\alpha\) on \(n\) is tight up to constant
factors.

\section{Conclusion and Future Work}
\label{sec:conclusion}
In stochastic \OMMT, matching lower bounds for arbitrary algorithms
and improved \(n,\Delta\) dependence remain open, as does the necessity
of the \(\sqrt T\) term for multiple minimizers.
In adversarial \OMMT, open directions include improving
\(O(n^2\sqrt T)\) regret and finding polynomial-time algorithms
with \(\alpha=O(\log n)\). LP relaxation and rounding of our MIP
may help if surrogate loss can be controlled.

\begin{credits}
\subsubsection{\ackname}
This work was supported by 
JSPS KAKENHI Grant Numbers JP23K24905, JP23K28038 and JP24H00685,
respectively.
\end{credits}

\bibliographystyle{splncs04}
\bibliography{references}

@inproceedings{Ailon2014,
  author    = {Nir Ailon},
  title     = {Improved Bounds for Online Learning over the Permutahedron and Other Ranking Polytopes},
  booktitle = {Proceedings of the 17th International Conference on Artificial Intelligence and Statistics},
  series    = {Proceedings of Machine Learning Research},
  volume    = {33},
  pages     = {29--37},
  year      = {2014},
  url       = {https://proceedings.mlr.press/v33/ailon14.html}
}

@article{Azuma1967,
  author  = {Kazuoki Azuma},
  title   = {Weighted Sums of Certain Dependent Random Variables},
  journal = {Tohoku Mathematical Journal, Second Series},
  volume  = {19},
  number  = {3},
  pages   = {357--367},
  year    = {1967},
  doi     = {10.2748/tmj/1178243286}
}

@book{CesaBianchiLugosi2006,
  author    = {Nicol{\`o} Cesa-Bianchi and G{\'a}bor Lugosi},
  title     = {Prediction, Learning, and Games},
  publisher = {Cambridge University Press},
  year      = {2006},
  doi       = {10.1017/CBO9780511546921}
}

@inproceedings{EvenDarKleinbergMannorMansour2009,
  author    = {Eyal Even-Dar and Robert Kleinberg and Shie Mannor and Yishay Mansour},
  title     = {Online Learning for Global Cost Functions},
  booktitle = {Proceedings of the 22nd Annual Conference on Learning Theory},
  year      = {2009}
}

@inproceedings{EvenDarMannorMansour2010,
  author    = {Eyal Even-Dar and Shie Mannor and Yishay Mansour},
  title     = {Learning with Global Cost in Stochastic Environments},
  booktitle = {Proceedings of the 23rd Annual Conference on Learning Theory},
  pages     = {80--92},
  publisher = {Omnipress},
  year      = {2010}
}

@inproceedings{Fujita2015,
  author    = {Takahiro Fujita and Kohei Hatano and Shuji Kijima and Eiji Takimoto},
  title     = {Online Linear Optimization for Job Scheduling Under Precedence Constraints},
  booktitle = {Proceedings of the 26th International Conference on Algorithmic Learning Theory},
  series    = {Lecture Notes in Computer Science},
  volume    = {9355},
  pages     = {332--346},
  publisher = {Springer},
  year      = {2015},
  doi       = {10.1007/978-3-319-24486-0_22}
}

@article{Hazan2016,
  author  = {Elad Hazan},
  title   = {Introduction to Online Convex Optimization},
  journal = {Foundations and Trends in Optimization},
  volume  = {2},
  number  = {3--4},
  pages   = {157--325},
  year    = {2016},
  doi     = {10.1561/2400000013}
}

@article{Hoeffding1963,
  author  = {Wassily Hoeffding},
  title   = {Probability Inequalities for Sums of Bounded Random Variables},
  journal = {Journal of the American Statistical Association},
  volume  = {58},
  number  = {301},
  pages   = {13--30},
  year    = {1963},
  doi     = {10.1080/01621459.1963.10500830}
}

@article{KalaiVempala2005,
  author  = {Adam Tauman Kalai and Santosh S. Vempala},
  title   = {Efficient Algorithms for Online Decision Problems},
  journal = {Journal of Computer and System Sciences},
  volume  = {71},
  number  = {3},
  pages   = {291--307},
  year    = {2005},
  doi     = {10.1016/j.jcss.2004.10.016}
}

@inproceedings{KesselheimSingla2020,
  author    = {Thomas Kesselheim and Sahil Singla},
  title     = {Online Learning with Vector Costs and Bandits with Knapsacks},
  booktitle = {Proceedings of the 33rd Conference on Learning Theory},
  series    = {Proceedings of Machine Learning Research},
  volume    = {125},
  pages     = {2286--2305},
  year      = {2020},
  url       = {https://proceedings.mlr.press/v125/kesselheim20a.html}
}

@article{Lawler1973,
  author  = {Eugene L. Lawler},
  title   = {Optimal Sequencing of a Single Machine Subject to Precedence Constraints},
  journal = {Management Science},
  volume  = {19},
  number  = {5},
  pages   = {544--546},
  year    = {1973},
  doi     = {10.1287/mnsc.19.5.544}
}

@article{LittlestoneWarmuth1994,
  author  = {Nick Littlestone and Manfred K. Warmuth},
  title   = {The Weighted Majority Algorithm},
  journal = {Information and Computation},
  volume  = {108},
  number  = {2},
  pages   = {212--261},
  year    = {1994},
  doi     = {10.1006/inco.1994.1009}
}

@inproceedings{LiuHatanoTakimoto2021,
  author    = {Yaxiong Liu and Kohei Hatano and Eiji Takimoto},
  title     = {Improved Algorithms for Online Load Balancing},
  booktitle = {{SOFSEM} 2021: Theory and Practice of Computer Science},
  series    = {Lecture Notes in Computer Science},
  volume    = {12607},
  pages     = {203--217},
  publisher = {Springer},
  year      = {2021},
  doi       = {10.1007/978-3-030-67731-2_15}
}

@book{Pinedo2016,
  author    = {Michael L. Pinedo},
  title     = {Scheduling: Theory, Algorithms, and Systems},
  edition   = {5},
  publisher = {Springer},
  year      = {2016},
  doi       = {10.1007/978-3-319-26580-3}
}

@inproceedings{Suehiro2012,
  author    = {Daiki Suehiro and Kohei Hatano and Shuji Kijima and Eiji Takimoto and Kiyohito Nagano},
  title     = {Online Prediction under Submodular Constraints},
  booktitle = {Proceedings of the 23rd International Conference on Algorithmic Learning Theory},
  series    = {Lecture Notes in Computer Science},
  volume    = {7568},
  pages     = {260--274},
  publisher = {Springer},
  year      = {2012},
  doi       = {10.1007/978-3-642-34106-9_22}
}

@inproceedings{SuggalaNetrapalli2020,
  author    = {Arun Sai Suggala and Praneeth Netrapalli},
  title     = {Online Non-Convex Learning: Following the Perturbed Leader is Optimal},
  booktitle = {Proceedings of the 31st International Conference on Algorithmic Learning Theory},
  series    = {Proceedings of Machine Learning Research},
  volume    = {117},
  pages     = {845--861},
  year      = {2020},
  url       = {https://proceedings.mlr.press/v117/suggala20a.html}
}

@inproceedings{Yasutake2011,
  author    = {Shota Yasutake and Kohei Hatano and Shuji Kijima and Eiji Takimoto and Masayuki Takeda},
  title     = {Online Linear Optimization over Permutations},
  booktitle = {Algorithms and Computation -- 22nd International Symposium, {ISAAC} 2011},
  series    = {Lecture Notes in Computer Science},
  volume    = {7074},
  pages     = {534--543},
  publisher = {Springer},
  year      = {2011},
  doi       = {10.1007/978-3-642-25591-5_55}
}

\ifwithappendix
\appendix
\section{Omitted Proofs}
\label{app:omitted-proofs}

This part collects proofs that are omitted from the main text.

\subsection{Proof of \Cref{prop:cumulative-lcl}}
\label{app:proof-cumulative-lcl}

\begin{proof}
Fix a round \(t\ge2\) and let \(m:=t-1\).
For a nonempty set \(S\subseteq[n]\), define
\[
  P_s(S):=\sum_{j\in S}p_j^{(s)},
  \qquad
  H_i^{[m]}(S)
  :=
  \sum_{s=1}^{m}
  \bigl(P_s(S)-d_i^{(s)}\bigr)_+,
  \qquad i\in S.
\]
Thus, \(H_i^{[m]}(S)/n\) is the cumulative loss of job \(i\) if it is
placed last among the jobs in \(S\).

For an ordering \(\pi\) of \(S\), let
\(
  Q_i^\pi
  :=
  \{j\in S:\pi(j)\le\pi(i)\}
\)
denote the set of jobs scheduled no later than \(i\).
Then
\[
  \sum_{s=1}^{m}\ell_i^{(s)}(\pi)
  =
  \frac{1}{n}H_i^{[m]}(Q_i^\pi).
\]
Hence, up to the common factor \(1/n\), the cumulative
\(\ell_\infty\)-objective on \(S\) is
\[
  F_S(\pi)
  :=
  \max_{i\in S}H_i^{[m]}(Q_i^\pi).
\]

The key property is that \(H_i^{[m]}\) is monotone with respect to set
inclusion. Indeed, if \(i\in S\subseteq S'\), then
\(P_s(S)\le P_s(S')\) for every \(s\in[m]\), and therefore
\[
  H_i^{[m]}(S)\le H_i^{[m]}(S').
\]

Now let
\(
  i^\star
  \in
  \operatorname*{arg\,min}_{i\in S}
  H_i^{[m]}(S)
\)
be the job selected by Extended LCL for the last remaining position.
We show that some optimal ordering of \(S\) has \(i^\star\) last.

Let \(\pi\) be an optimal ordering of \(S\), and let \(k\) be its last
job. If \(k=i^\star\), the claim is immediate. Otherwise, let \(\pi'\)
be obtained from \(\pi\) by moving \(i^\star\) to the last position
while preserving the relative order of all other jobs.
For every \(i\neq i^\star\),
\(Q_i^{\pi'}\subseteq Q_i^\pi\), so
\[
  H_i^{[m]}(Q_i^{\pi'})
  \le
  H_i^{[m]}(Q_i^\pi)
  \le
  F_S(\pi).
\]
For \(i^\star\), using
\(Q_{i^\star}^{\pi'}=Q_k^\pi=S\) and the choice of \(i^\star\),
\[
  H_{i^\star}^{[m]}(Q_{i^\star}^{\pi'})
  =
  H_{i^\star}^{[m]}(S)
  \le
  H_k^{[m]}(S)
  =
  H_k^{[m]}(Q_k^\pi)
  \le
  F_S(\pi).
\]
Therefore \(F_S(\pi')\le F_S(\pi)\), and optimality of \(\pi\) implies
that \(\pi'\) is also optimal. Thus, the last-position choice made by
Extended LCL is consistent with an optimal ordering.

After fixing \(i^\star\) in the last position, the value
\(H_{i^\star}^{[m]}(S)\) is fixed, while the losses of the remaining
jobs depend only on their relative ordering on
\(S\setminus\{i^\star\}\).
Applying the same argument recursively shows that every backward
choice of Extended LCL is consistent with an optimal ordering.
Starting from \(S=[n]\), the permutation \(\sigma^{(t)}\) returned by
Extended LCL therefore satisfies
\(
  \sigma^{(t)}
  \in
  \operatorname*{arg\,min}_{\sigma\in\Sn}
  \left\|
    \sum_{s=1}^{t-1}\bell^{(s)}(\sigma)
  \right\|_\infty .
\)

It remains to bound the computational cost.
For each remaining set \(S\), the \(m\) values \(P_s(S)\) can be
maintained throughout the backward construction.
Given these values, evaluating \(H_i^{[m]}(S)\) for one candidate
\(i\in S\) takes \(O(m)\) time.
Thus, if \(|S|=r\), selecting the next job takes \(O(mr)\) time, and
the total running time is
\(
  O\!\left(
    m\sum_{r=1}^{n}r
  \right)
  =
  O(mn^2).
\)
Updating all \(P_s(S)\) after a job is removed takes \(O(m)\) time per
iteration and is absorbed in this bound.

Storing the \(m\) observed processing-time and due-date vectors uses
\(O(mn)\) space; the maintained values \(P_s(S)\) and the permutation
require only \(O(m+n)\) additional space.
Since \(m=t-1\), the total complexity at round \(t\) is
\(O(tn^2)\) time and \(O(tn)\) space.
\end{proof}

\subsection{Proof of \Cref{lem:unique-ftl-stabilization}}
\label{app:proof-unique-ftl-stabilization}

\begin{proof}
Fix \(t\ge2\) and let \(m:=t-1\).
For \(\sigma\in\Sn\), define
\[
  \widehat L_m(\sigma)
  :=
  \left\|
    \frac{1}{m}
    \sum_{s=1}^{m}\bell^{(s)}(\sigma)
  \right\|_\infty .
\]
Since multiplying the objective by \(1/m\) does not change its
minimizers, \(\sigma^{(t)}\) minimizes \(\widehat L_m\).

Consider the event
\[
  \mathcal E_m
  :=
  \left\{
    \sup_{\sigma\in\Sn}
    \left\|
      \frac{1}{m}
      \sum_{s=1}^{m}\bell^{(s)}(\sigma)
      -
      \bm\mu(\sigma)
    \right\|_\infty
    \le
    \frac{\Delta}{4}
  \right\}.
\]
On \(\mathcal E_m\), the reverse triangle inequality gives
\[
  \left|
    \widehat L_m(\sigma)
    -
    \|\bm\mu(\sigma)\|_\infty
  \right|
  \le
  \frac{\Delta}{4}
  \qquad
  \text{for every }\sigma\in\Sn .
\]
Hence, for every \(\sigma\neq\sigma^\star\),
\[
  \widehat L_m(\sigma)
  \ge
  \|\bm\mu(\sigma)\|_\infty-\frac{\Delta}{4}
  \ge
  \|\bm\mu(\sigma^\star)\|_\infty+\frac{3\Delta}{4},
\]
whereas
\[
  \widehat L_m(\sigma^\star)
  \le
  \|\bm\mu(\sigma^\star)\|_\infty+\frac{\Delta}{4}.
\]
Therefore, on \(\mathcal E_m\), the unique minimizer of
\(\widehat L_m\) is \(\sigma^\star\). Consequently,
\(
  \{\sigma^{(t)}\neq\sigma^\star\}
  \subseteq
  \mathcal E_m^c .
\)

For a fixed \((i,\sigma)\in[n]\times\Sn\), Hoeffding's
inequality~\cite{Hoeffding1963} yields
\[
  \Prob\!\left(
    \left|
      \frac{1}{m}
      \sum_{s=1}^{m}\ell_i^{(s)}(\sigma)
      -
      \mu_i(\sigma)
    \right|
    >
    \frac{\Delta}{4}
  \right)
  \le
  2\exp\!\left(
    -\frac{m\Delta^2}{8}
  \right).
\]
A union bound over the \(nn!\) pairs \((i,\sigma)\) therefore gives
\[
  \Prob(\sigma^{(t)}\neq\sigma^\star)
  \le
  2nn!
  \exp\!\left(
    -\frac{(t-1)\Delta^2}{8}
  \right),
\]
which proves the first claim.

We next bound \(\E[N_T]\).
Set
\(
  a:=2nn!
\)
and
\(
  b:=\Delta^2/8
\).
Since the first-round permutation is arbitrary,
\[
  \E[N_T]
  \le
  1+
  \sum_{m=1}^{T-1}
  \min\{1,ae^{-bm}\}
  \le
  1+
  \sum_{m=1}^{\infty}
  \min\{1,ae^{-bm}\}.
\]
Let
\(
  m_0:=\lceil (\log a)/b\rceil
\).
Then
\[
  \sum_{m=1}^{\infty}\min\{1,ae^{-bm}\}
  \le
  m_0+
  \frac{ae^{-b(m_0+1)}}{1-e^{-b}}
  \le
  m_0+\frac{1}{1-e^{-b}}.
\]
Because the loss vectors lie in \([0,1]^n\), we have
\(0<\Delta\le1\), and hence \(b\in(0,1]\).
Using \(1-e^{-b}\ge b/2\),
\[
  \E[N_T]
  =
  O\!\left(
    \frac{\log a}{b}
  \right)
  =
  O\!\left(
    \frac{n\log n}{\Delta^2}
  \right).
\]

Finally, for every integer \(\tau\ge1\),
\[
  \Prob\!\left(
    \exists\,t>\tau:
    \sigma^{(t)}\neq\sigma^\star
  \right)
  \le
  \sum_{t=\tau+1}^{\infty}
  ae^{-b(t-1)}
  =
  \frac{ae^{-b\tau}}{1-e^{-b}}.
\]
Define
\[
  \tau_\Delta(\delta)
  :=
  \left\lceil
    \frac{1}{b}
    \left(
      \log\frac{a}{\delta}
      +
      \log\frac{1}{1-e^{-b}}
    \right)
  \right\rceil .
\]
Then, with probability at least \(1-\delta\),
\(
  N_T
  \le
  \tau_\Delta(\delta).
\)
Since \(1-e^{-b}\ge b/2\) and \(b=\Delta^2/8\),
\[
  \log\frac{1}{1-e^{-b}}
  =
  O\!\left(
    \log\frac{1}{\Delta}
  \right).
\]
Together with
\(
  \log a=\log(2nn!)=O(n\log n)
\),
this yields
\[
  N_T
  =
  O\!\left(
    \frac{
      n\log n
      +
      \log(1/\delta)
      +
      \log(1/\Delta)
    }{\Delta^2}
  \right)
\]
with probability at least \(1-\delta\).
\end{proof}

\subsection{Proof of \Cref{lem:unique-benchmark}}
\label{app:proof-unique-benchmark}

\begin{proof}
By the definition of \(\OPT_T\),
\(
  B_T\ge0
\).
Moreover, since all loss coordinates lie in \([0,1]\),
\(
  B_T\le T
\).

Consider the event
\[
  \mathcal E_T
  :=
  \left\{
    \sup_{\sigma\in\Sn}
    \left\|
      \frac{1}{T}
      \sum_{t=1}^{T}\bell^{(t)}(\sigma)
      -
      \bm\mu(\sigma)
    \right\|_\infty
    \le
    \frac{\Delta}{4}
  \right\}.
\]
As in the proof of \Cref{lem:unique-ftl-stabilization},
on \(\mathcal E_T\), \(\sigma^\star\) is the unique minimizer of
\(
  \left\|
    \sum_{t=1}^{T}\bell^{(t)}(\sigma)
  \right\|_\infty
\).
Hence \(B_T=0\) on \(\mathcal E_T\), and therefore
\(
  \{B_T>0\}
  \subseteq
  \mathcal E_T^c.
\)

For each fixed \((i,\sigma)\in[n]\times\Sn\), Hoeffding's
inequality~\cite{Hoeffding1963} gives
\[
  \Prob\!\left(
    \left|
      \frac{1}{T}
      \sum_{t=1}^{T}\ell_i^{(t)}(\sigma)
      -
      \mu_i(\sigma)
    \right|
    >
    \frac{\Delta}{4}
  \right)
  \le
  2\exp\!\left(-\frac{T\Delta^2}{8}\right).
\]
A union bound over all \(nn!\) pairs gives
\[
  \Prob(B_T>0)
  \le
  \Prob(\mathcal E_T^c)
  \le
  2nn!\exp\!\left(-\frac{T\Delta^2}{8}\right).
\]

Let
\(
  a:=2nn!
\)
and
\(
  b:=\Delta^2/8
\).
Since \(0\le B_T\le T\),
\(
  \E[B_T]
  \le
  T\min\{1,ae^{-bT}\}.
\)
Let
\(
  T_0:=(1+\log a)/b
\).
For \(T\le T_0\), we have \(\E[B_T]\le T_0\).
For \(T>T_0\), since \(x e^{-bx}\) is decreasing for
\(x\ge T_0>1/b\),
\[
  \E[B_T]
  \le
  aTe^{-bT}
  \le
  aT_0e^{-bT_0}
  \le
  T_0.
\]
Thus
\[
  \E[B_T]
  =
  O\!\left(
    \frac{1+\log a}{b}
  \right)
  =
  O\!\left(
    \frac{n\log n}{\Delta^2}
  \right).
\]

Finally, fix \(\delta\in(0,1)\).
If
\(
  T\ge b^{-1}\log(a/\delta)
\),
then
\(
  \Prob(B_T>0)\le\delta
\),
so \(B_T=0\) with probability at least \(1-\delta\).
Otherwise,
\(
  B_T
  \le
  T
  <
  \frac{1}{b}\log\frac{a}{\delta}.
\)
Hence, for every \(T\), with probability at least \(1-\delta\),
\[
  B_T
  =
  O\!\left(
    \frac{
      n\log n+\log(1/\delta)
    }{\Delta^2}
  \right).
\]
\end{proof}

\subsection{Proof of \Cref{prop:max-btl-counterexample}}
\label{app:proof-max-btl-counterexample}

\begin{proof}
For \(x\in\mathcal X\), define its cumulative vector loss up to round
\(t\) by
\[
  \bm L^{(t)}(x)
  :=
  \sum_{s=1}^{t}\bell^{(s)}(x).
\]

Let
\(
  \mathcal X=\{c,a_1,\ldots,a_T\},
\)
where \(c\) is the fixed comparator and \(a_t\) will be the unique
\(\ell_\infty\)-leader used at round \(t+1\).
Fix a constant \(\delta\in(0,1)\) and define
\(
  A_t:=1+t\delta/T
\),
\(t=1,\ldots,T\).

For the comparator \(c\), define
\[
  \bell^{(1)}(c):=(0,A_1),
  \qquad
  \bell^{(t)}(c):=(0,A_t-A_{t-1}),
  \quad t\ge2.
\]
Then
\(
  \bm L^{(t)}(c)=(0,A_t)
\),
so
\(
  \|\bm L^{(t)}(c)\|_\infty=A_t
\)
and
\(
  \|\bm L^{(T)}(c)\|_\infty=1+\delta
\).

Choose constants \(\theta,\lambda>0\) such that
\[
  \max\{\theta,\lambda\}<A_1,
  \qquad
  \theta+\lambda>A_1.
\]
For example, for sufficiently small \(\delta\), one may take
\(\theta=\lambda=3/4\).

Define
\(
  \bell^{(1)}(a_1):=(\theta,\lambda)
\).
Since
\(
  \|\bm L^{(1)}(a_1)\|_\infty
  =
  \max\{\theta,\lambda\}
  <
  A_1
  =
  \|\bm L^{(1)}(c)\|_\infty
\),
\(a_1\) is strictly better than \(c\) in the leader computation for
round \(2\).
Assign sufficiently large losses to \(a_1\) in subsequent rounds so
that it is never selected again.

For each \(r=2,\ldots,T\), choose
\(
  0<\varepsilon_r<A_r-A_{r-1}=\delta/T
\).
For every \(t<r\), define
\[
  \bell^{(t)}(a_r)
  :=
  \left(
    0,
    A_t-A_{t-1}+\frac{\varepsilon_r}{r-1}
  \right).
\]
Then
\[
  \bm L^{(t)}(a_r)
  =
  \left(
    0,
    A_t+\frac{t}{r-1}\varepsilon_r
  \right),
  \qquad t<r,
\]
and hence
\[
  \|\bm L^{(t)}(a_r)\|_\infty
  >
  A_t
  =
  \|\bm L^{(t)}(c)\|_\infty.
\]
Thus, before round \(r+1\), \(a_r\) is never preferred to \(c\) in any
earlier leader computation.

At round \(r\), set
\(
  \bell^{(r)}(a_r):=(\theta,0).
\)
Then
\(
  \bm L^{(r)}(a_r)
  =
  (\theta,A_{r-1}+\varepsilon_r).
\)
Since
\(
  \theta<A_1\le A_r
\)
and
\(
  A_{r-1}+\varepsilon_r<A_r
\),
we obtain
\[
  \|\bm L^{(r)}(a_r)\|_\infty
  <
  A_r
  =
  \|\bm L^{(r)}(c)\|_\infty.
\]
Hence \(a_r\) is strictly better than \(c\) in the leader computation
for round \(r+1\).
For \(t>r\), assign sufficiently large losses to \(a_r\) so that it is
never selected again.

Now fix \(t\in[T]\).
All actions \(a_r\) with \(r<t\) have already been excluded by their
subsequent large losses, while every \(a_r\) with \(r>t\) is strictly
worse than \(c\) in the leader computation for round \(t+1\).
Since \(a_t\) is strictly better than \(c\), it is the unique
\(\ell_\infty\)-leader used at round \(t+1\).

Therefore,
\[
  \sum_{t=1}^{T}
  \bell^{(t)}(x_\infty^{(t+1)})
  =
  \sum_{t=1}^{T}
  \bell^{(t)}(a_t)
  =
  (T\theta,\lambda),
\]
and hence
\[
  \left\|
    \sum_{t=1}^{T}
    \bell^{(t)}(x_\infty^{(t+1)})
  \right\|_\infty
  \ge
  T\theta.
\]
On the other hand,
\[
  \left\|
    \sum_{t=1}^{T}\bell^{(t)}(c)
  \right\|_\infty
  =
  1+\delta.
\]
Thus, any factor \(\alpha\) satisfying
\eqref{eq:extended-ftl-btl} must obey
\[
  \alpha
  \ge
  \frac{T\theta}{1+\delta}
  =
  \Omega(T).
\]
\end{proof}

\section{Stochastic \OMMT with Multiple Minimizers}
\label{app:stochastic-multiple-optima}

We extend the stochastic analysis to the case in which
\(
  \|\bm\mu(\sigma)\|_\infty
\)
have multiple minimizers.
Let
\(
  \cS^\star
  :=
  \operatorname*{arg\,min}_{\sigma\in\Sn}
  \|\bm\mu(\sigma)\|_\infty,
\)
and
\(
  L^\star
  :=
  \min_{\sigma\in\Sn}
  \|\bm\mu(\sigma)\|_\infty .
\)
Assume that \(\cS^\star\neq\Sn\), and define
\[
  \Delta
  :=
  \min_{\sigma\notin\cS^\star}
  \left(
    \|\bm\mu(\sigma)\|_\infty-L^\star
  \right).
\]
We also define the population benchmark
\(
  \OPT_T^{\mathrm{mean}}
  :=
  T L^\star .
\)
This is the population counterpart of the realized hindsight benchmark
\(\OPT_T\), analogous to the expected-loss benchmark used in standard
pseudo-regret analyses for stochastic online learning.

Fix an arbitrary
\(
  \sigma_{\mathrm{mean}}^\star\in\cS^\star
\).
For each round \(t\), define
\[
  \sigma_t^\star
  :=
  \begin{cases}
    \sigma^{(t)},
      & \text{if }\sigma^{(t)}\in\cS^\star,\\
    \sigma_{\mathrm{mean}}^\star,
      & \text{otherwise}.
  \end{cases}
\]
Thus, \(\sigma_t^\star\in\cS^\star\) for every \(t\), and
\(\sigma_t^\star=\sigma^{(t)}\) whenever
\(\sigma^{(t)}\in\cS^\star\).

Define
\(
  D_T
  :=
  \sup_{\sigma\in\Sn}
  \left\|
    \sum_{t=1}^{T}\bell^{(t)}(\sigma)
    -
    T\bm\mu(\sigma)
  \right\|_\infty .
\)
Comparing the permutation selected by FTL in each round with
\(\sigma_t^\star\), the regret can be decomposed as
\begin{equation}
\label{eq:multi-ftl-decomposition}
\begin{aligned}
  \Reg_T^{\FTL}
  &=
  \left\|
    \sum_{t=1}^{T}\bell^{(t)}(\sigma^{(t)})
  \right\|_\infty
  -
  \min_{\sigma\in\Sn}
  \left\|
    \sum_{t=1}^{T}
    \bell^{(t)}(\sigma)
  \right\|_\infty
  \\
  &\le
  \left\|
    \sum_{t=1}^{T}\bell^{(t)}(\sigma_t^\star)
  \right\|_\infty
  +
  \underbrace{
    \sum_{t=1}^{T}
    \mathbf 1\{\sigma^{(t)}\notin\cS^\star\}
  }_{=:N_T}
  -
  \OPT_T
  && \text{(a)}
  \\
  &\le
  \left\|
    \sum_{t=1}^{T}\bm\mu(\sigma_t^\star)
  \right\|_\infty
  +
  \underbrace{
    \left\|
      \sum_{t=1}^{T}
      \bigl(
        \bell^{(t)}(\sigma_t^\star)
        -
        \bm\mu(\sigma_t^\star)
      \bigr)
    \right\|_\infty
  }_{=:M_T}
  +
  N_T
  -
  \OPT_T
  && \text{(b)}
  \\
  &\le
  \OPT_T^{\mathrm{mean}}
  +
  M_T+N_T-\OPT_T
  && \text{(c)}
  \\
  &\le
  D_T+M_T+N_T
  && \text{(d)}.
\end{aligned}
\end{equation}
For (a), since all loss coordinates lie in \([0,1]\),
\(
  \|\bell^{(t)}(\sigma^{(t)})
  -
  \bell^{(t)}(\sigma_t^\star)\|_\infty
  \le
  \mathbf 1\{\sigma^{(t)}\notin\cS^\star\}
\).
For (b), we apply the triangle inequality.
For (c), since \(\sigma_t^\star\in\cS^\star\) for every \(t\),
\[
  \left\|
    \sum_{t=1}^{T}\bm\mu(\sigma_t^\star)
  \right\|_\infty
  \le
  \sum_{t=1}^{T}
  \|\bm\mu(\sigma_t^\star)\|_\infty
  =
  T L^\star
  =
  \OPT_T^{\mathrm{mean}}.
\]
Finally, the definition of \(D_T\) implies
\(
  \OPT_T
  \ge
  \OPT_T^{\mathrm{mean}}-D_T
\),
which gives (d).

The following lemmas bound the three terms in
\eqref{eq:multi-ftl-decomposition}.

\begin{lemma}
\label{lem:multi-benchmark-concentration}

\(
  \E[D_T]
  =
  O(\sqrt{Tn\log n})
\).
Moreover, for every \(\delta\in(0,1)\), with probability at least
\(1-\delta\),
\(
  D_T
  =
  O(\sqrt{T(n\log n+\log(1/\delta))})
\).

\end{lemma}

\begin{proof}
Fix any \((i,\sigma)\in[n]\times\Sn\).
Since
\(
  \ell_i^{(1)}(\sigma),\ldots,\ell_i^{(T)}(\sigma)
\)
are i.i.d.\ random variables in \([0,1]\) with mean
\(\mu_i(\sigma)\), Hoeffding's inequality~\cite{Hoeffding1963}
implies that, for every \(r>0\),
\[
  \Prob\!\left(
    \left|
      \sum_{t=1}^{T}\ell_i^{(t)}(\sigma)
      -
      T\mu_i(\sigma)
    \right|
    \ge r
  \right)
  \le
  2\exp\!\left(-\frac{2r^2}{T}\right).
\]
A union bound over all \(nn!\) pairs \((i,\sigma)\) therefore gives
\begin{equation}
\label{eq:multi-DT-tail}
  \Prob(D_T\ge r)
  \le
  2nn!\exp\!\left(-\frac{2r^2}{T}\right).
\end{equation}

For the expectation bound, let
\(
  a:=2nn!
\)
and
\(
  r_0:=\sqrt{(T/2)\log a}
\).
Using \eqref{eq:multi-DT-tail} and
\(
  \E[D_T]
  =
  \int_0^\infty\Prob(D_T\ge r)\,dr
\),
we obtain
\[
  \E[D_T]
  \le
  r_0+
  a\int_{r_0}^{\infty}
  \exp\!\left(-\frac{2r^2}{T}\right)\,dr .
\]
Using the standard Gaussian tail bound
\[
  \int_x^\infty
  \exp\!\left(-\frac{2r^2}{T}\right)\,dr
  \le
  \frac{T}{4x}
  \exp\!\left(-\frac{2x^2}{T}\right),
  \qquad x>0,
\]
we obtain
\[
  \E[D_T]
  \le
  r_0+\frac{T}{4r_0}
  =
  O(\sqrt{Tn\log n}).
\]

Finally, taking
\(
  r=\sqrt{(T/2)\log(2nn!/\delta)}
\)
in \eqref{eq:multi-DT-tail}, we obtain, with probability at least
\(1-\delta\),
\[
  D_T
  \le
  \sqrt{
    \frac{T}{2}
    \log\frac{2nn!}{\delta}
  }
  =
  O\!\left(
    \sqrt{T(n\log n+\log(1/\delta))}
  \right).
\]
\end{proof}

\begin{lemma}
\label{lem:multi-path-concentration}

\(
  \E[M_T]
  =
  O(\sqrt{T\log n})
\).
Moreover, for every \(\delta\in(0,1)\), with probability at least
\(1-\delta\),
\(
  M_T
  =
  O(\sqrt{T(\log n+\log(1/\delta))})
\).

\end{lemma}

\begin{proof}
Let
\(
  \mathcal H_{t-1}
  :=
  \sigma\!\left(
    (\bm p^{(1)},\bm d^{(1)}),\ldots,
    (\bm p^{(t-1)},\bm d^{(t-1)})
  \right)
\)
be the sigma-field generated by the instances observed before round
\(t\).
By construction, \(\sigma_t^\star\) is
\(\mathcal H_{t-1}\)-measurable.
Since the round-\(t\) instance is independent of
\(\mathcal H_{t-1}\),
\[
  \E\!\left[
    \bell^{(t)}(\sigma_t^\star)
    \,\middle|\,
    \mathcal H_{t-1}
  \right]
  =
  \bm\mu(\sigma_t^\star).
\]

For each \(i\in[n]\), define
\(
  X_{i,t}
  :=
  \ell_i^{(t)}(\sigma_t^\star)
  -
  \mu_i(\sigma_t^\star).
\)
Then \((X_{i,t})_{t=1}^{T}\) is a martingale difference sequence with
respect to \((\mathcal H_t)_{t=0}^{T}\).
Moreover, since
\(
  \ell_i^{(t)}(\sigma_t^\star),
  \mu_i(\sigma_t^\star)\in[0,1]
\),
we have \(|X_{i,t}|\le1\).

Hence, by the Azuma--Hoeffding inequality~\cite{Azuma1967}, for every
\(r>0\),
\[
  \Prob\!\left(
    \left|
      \sum_{t=1}^{T}X_{i,t}
    \right|
    \ge r
  \right)
  \le
  2\exp\!\left(
    -\frac{r^2}{2T}
  \right).
\]
Taking a union bound over the \(n\) coordinates gives
\begin{equation}
\label{eq:multi-MT-tail}
  \Prob(M_T\ge r)
  \le
  2n\exp\!\left(
    -\frac{r^2}{2T}
  \right).
\end{equation}

For the expectation bound, let
\(
  r_0:=\sqrt{2T\log(2n)}
\).
Using \eqref{eq:multi-MT-tail} and
\(
  \E[M_T]
  =
  \int_0^\infty\Prob(M_T\ge r)\,dr
\),
we obtain
\[
  \E[M_T]
  \le
  r_0+
  2n\int_{r_0}^{\infty}
  \exp\!\left(
    -\frac{r^2}{2T}
  \right)dr .
\]
Using the Gaussian tail bound
\[
  \int_x^\infty
  \exp\!\left(
    -\frac{r^2}{2T}
  \right)dr
  \le
  \frac{T}{x}
  \exp\!\left(
    -\frac{x^2}{2T}
  \right),
  \qquad x>0,
\]
we obtain
\[
  \E[M_T]
  \le
  r_0+\frac{T}{r_0}
  =
  O(\sqrt{T\log n}).
\]

Finally, taking
\(
  r=\sqrt{2T\log(2n/\delta)}
\)
in \eqref{eq:multi-MT-tail}, we obtain, with probability at least
\(1-\delta\),
\[
  M_T
  \le
  \sqrt{2T\log\frac{2n}{\delta}}
  =
  O\!\left(
    \sqrt{T(\log n+\log(1/\delta))}
  \right).
\]
\end{proof}

\begin{lemma}
\label{lem:multi-ftl-stabilization}
For every \(t\ge2\),
\(
  \Prob(\sigma^{(t)}\notin\cS^\star)
  \le
  2nn!\exp(-(t-1)\Delta^2/8)
\),
and
\(
  \E[N_T]
  =
  O(n\log n/\Delta^2)
\).
Moreover, for every \(\delta\in(0,1)\), with probability at least
\(1-\delta\),
\(
  N_T
  =
  O(
    (n\log n+\log(1/\delta)+\log(1/\Delta))/\Delta^2
  )
\).
\end{lemma}

\begin{proof}
Fix \(t\ge2\) and let \(m:=t-1\).
For \(\sigma\in\Sn\), define
\[
  \widehat L_m(\sigma)
  :=
  \left\|
    \frac{1}{m}
    \sum_{s=1}^{m}\bell^{(s)}(\sigma)
  \right\|_\infty,
  \qquad
  L(\sigma)
  :=
  \|\bm\mu(\sigma)\|_\infty .
\]
Since multiplying the objective by \(1/m\) does not change its
minimizers, \(\sigma^{(t)}\) minimizes \(\widehat L_m\).

Consider the event
\[
  \mathcal E_m
  :=
  \left\{
    \sup_{\sigma\in\Sn}
    \left\|
      \frac{1}{m}
      \sum_{s=1}^{m}\bell^{(s)}(\sigma)
      -
      \bm\mu(\sigma)
    \right\|_\infty
    \le
    \frac{\Delta}{4}
  \right\}.
\]
On \(\mathcal E_m\), the reverse triangle inequality gives
\[
  \bigl|
    \widehat L_m(\sigma)-L(\sigma)
  \bigr|
  \le
  \frac{\Delta}{4}
  \qquad
  \text{for every }\sigma\in\Sn .
\]
Hence, for every \(\sigma\notin\cS^\star\),
\[
  \widehat L_m(\sigma)
  \ge
  L(\sigma)-\frac{\Delta}{4}
  \ge
  L^\star+\frac{3\Delta}{4},
\]
whereas every \(\sigma^\star\in\cS^\star\) satisfies
\[
  \widehat L_m(\sigma^\star)
  \le
  L^\star+\frac{\Delta}{4}.
\]
Therefore, on \(\mathcal E_m\), every minimizer of
\(\widehat L_m\) belongs to \(\cS^\star\), and
\(
  \{\sigma^{(t)}\notin\cS^\star\}
  \subseteq
  \mathcal E_m^c .
\)

For each fixed \((i,\sigma)\in[n]\times\Sn\), Hoeffding's
inequality~\cite{Hoeffding1963} gives
\[
  \Prob\!\left(
    \left|
      \frac{1}{m}
      \sum_{s=1}^{m}\ell_i^{(s)}(\sigma)
      -
      \mu_i(\sigma)
    \right|
    >
    \frac{\Delta}{4}
  \right)
  \le
  2\exp\!\left(
    -\frac{m\Delta^2}{8}
  \right).
\]
A union bound over the \(nn!\) pairs \((i,\sigma)\) therefore gives
\[
  \Prob(\sigma^{(t)}\notin\cS^\star)
  \le
  2nn!
  \exp\!\left(
    -\frac{(t-1)\Delta^2}{8}
  \right).
\]

Set
\(
  a:=2nn!
\)
and
\(
  b:=\Delta^2/8
\).
Since the first-round permutation is arbitrary,
\[
  \E[N_T]
  \le
  1+
  \sum_{m=1}^{\infty}
  \min\{1,ae^{-bm}\}.
\]
Let
\(
  m_0:=\lceil(\log a)/b\rceil
\).
Then
\[
  \sum_{m=1}^{\infty}\min\{1,ae^{-bm}\}
  \le
  m_0+\frac{1}{1-e^{-b}}.
\]
Since \(0<\Delta\le1\), we have \(1-e^{-b}\ge b/2\), and therefore
\[
  \E[N_T]
  =
  O\!\left(
    \frac{n\log n}{\Delta^2}
  \right).
\]

Finally, for every integer \(\tau\ge1\),
\[
  \Prob\!\left(
    \exists\,t>\tau:
    \sigma^{(t)}\notin\cS^\star
  \right)
  \le
  \frac{ae^{-b\tau}}{1-e^{-b}}.
\]
Taking
\[
  \tau_\Delta(\delta)
  :=
  \left\lceil
    \frac{1}{b}
    \left(
      \log\frac{a}{\delta}
      +
      \log\frac{1}{1-e^{-b}}
    \right)
  \right\rceil
\]
gives \(N_T\le\tau_\Delta(\delta)\) with probability at least
\(1-\delta\).
Using
\(
  \log a=O(n\log n)
\)
and
\(
  \log(1/(1-e^{-b}))
  =
  O(\log(1/\Delta))
\)
proves the stated high-probability bound.
\end{proof}

Combining the preceding lemmas with
\eqref{eq:multi-ftl-decomposition} gives the following regret bounds.

\begin{theorem}
\label{thm:multi-stoch-ftl-expectation}
For stochastic \OMMT with multiple minimizers, FTL satisfies
\[
  \E[\Reg_T^{\FTL}]
  =
  O\!\left(
    \sqrt{Tn\log n}
    +
    \frac{n\log n}{\Delta^2}
  \right).
\]
\end{theorem}

\begin{theorem}
\label{thm:multi-stoch-ftl-highprob}
For every \(\delta\in(0,1)\), FTL for stochastic \OMMT with 
multiple minimizers satisfies, with probability at least \(1-\delta\),
\[
  \Reg_T^{\FTL}
  =
  O\!\left(
    \sqrt{
      T\bigl(n\log n+\log(1/\delta)\bigr)
    }
    +
    \frac{
      n\log n+\log(1/\delta)+\log(1/\Delta)
    }{\Delta^2}
  \right).
\]
\end{theorem}

\section{Additional Observations on the Extended FTL--BTL Lemma}
\label{app:additional-observations}

This section collects several additional observations concerning
extensions of the FTL--BTL lemma to vector-valued losses.
These results do not provide a direct adversarial analysis for \OMMT,
but clarify how the behavior depends on the choice of norm and leader.

\subsection{Failure for Finite \texorpdfstring{\(p>1\)}{p>1}}
\label{app:finite-p-counterexample}

The \(\ell_\infty\)-norm counterexample in
\Cref{prop:max-btl-counterexample} is the case directly relevant to
\OMMT.
The same obstruction also appears for finite
\(p\in(1,\infty)\).

Let
\(
  x_p^{(t+1)}
  \in
  \operatorname*{arg\,min}_{x\in\mathcal X}
  \left\|
    \sum_{s=1}^{t}\bell^{(s)}(x)
  \right\|_p .
\)
The corresponding extension of the FTL--BTL lemma would require
a horizon-independent factor \(\alpha\) such that, for every fixed
\(u\in\mathcal X\),
\begin{equation}
\label{eq:p-norm-btl-question}
  \left\|
    \sum_{t=1}^{T}
    \bell^{(t)}(x_p^{(t+1)})
  \right\|_p
  \le
  \alpha
  \left\|
    \sum_{t=1}^{T}\bell^{(t)}(u)
  \right\|_p .
\end{equation}

\begin{proposition}
\label{prop:finite-p-btl-counterexample}
For every \(p\in(1,\infty)\), there exists a two-dimensional
nonnegative vector-loss instance with a unique \(\ell_p\)-leader at
each round such that any inequality of the form
\eqref{eq:p-norm-btl-question} requires
\[
  \alpha
  =
  \Omega\!\left(T^{1-1/p}\right).
\]
\end{proposition}

\begin{proof}
For \(x\in\mathcal X\), define
\(
  \bm L^{(t)}(x)
  :=
  \sum_{s=1}^{t}\bell^{(s)}(x).
\)
Let
\(
  \mathcal X=\{c,a_1,\ldots,a_T\},
\)
where \(c\) is the fixed comparator and \(a_t\) will be the unique
\(\ell_p\)-leader used at round \(t+1\).
Set
\(
  A^{(t)}:=t^{1/p}
\)
and \(A^{(0)}:=0\).
Choose constants \(\theta,\lambda>0\) such that
\[
  2^{1/p-1}<\theta<1,
  \qquad
  1-\theta<\lambda<(1-\theta^p)^{1/p}.
\]

For the comparator \(c\), define
\(
  \bell^{(t)}(c)
  :=
  \bigl(0,A^{(t)}-A^{(t-1)}\bigr).
\)
Then
\(
  \bm L^{(t)}(c)=(0,A^{(t)})
\),
and hence
\(
  \|\bm L^{(t)}(c)\|_p
  =
  A^{(t)}
  =
  t^{1/p}.
\)

For \(a_1\), set
\(
  \bell^{(1)}(a_1):=(\theta,\lambda)
\).
Since
\(
  \|\bm L^{(1)}(a_1)\|_p
  =
  (\theta^p+\lambda^p)^{1/p}
  <
  1
  =
  \|\bm L^{(1)}(c)\|_p,
\)
\(a_1\) is strictly better than \(c\) in the leader computation for
round \(2\).
Assign sufficiently large losses to \(a_1\) in subsequent rounds so
that it is never selected again.

For each \(r=2,\ldots,T\), choose \(\varepsilon_r>0\) sufficiently
small that
\(
  \theta^p+
  \bigl(A^{(r-1)}+\varepsilon_r\bigr)^p
  <
  \bigl(A^{(r)}\bigr)^p.
\)
For every \(t<r\), define
\[
  \bell^{(t)}(a_r)
  :=
  \left(
    0,
    A^{(t)}-A^{(t-1)}
    +
    \frac{\varepsilon_r}{r-1}
  \right).
\]
Then
\[
  \bm L^{(t)}(a_r)
  =
  \left(
    0,
    A^{(t)}
    +
    \frac{t}{r-1}\varepsilon_r
  \right),
  \qquad t<r,
\]
so
\(
  \|\bm L^{(t)}(a_r)\|_p
  >
  \|\bm L^{(t)}(c)\|_p
\).
Thus \(a_r\) is strictly worse than \(c\) in every leader computation
before round \(r+1\).

At round \(r\), set
\(
  \bell^{(r)}(a_r):=(\theta,0).
\)
Then
\[
  \|\bm L^{(r)}(a_r)\|_p
  =
  \left(
    \theta^p+
    \bigl(A^{(r-1)}+\varepsilon_r\bigr)^p
  \right)^{1/p}
  <
  A^{(r)}
  =
  \|\bm L^{(r)}(c)\|_p.
\]
Hence \(a_r\) is strictly better than \(c\) in the leader computation
for round \(r+1\).
Assign sufficiently large losses to \(a_r\) in subsequent rounds so
that it is never selected again.

As in the \(\ell_\infty\) construction, for every \(t\in[T]\), all
actions \(a_r\) with \(r<t\) have already been excluded by their
subsequent large losses, while every \(a_r\) with \(r>t\) is strictly
worse than \(c\).
Therefore, \(a_t\) is the unique \(\ell_p\)-leader used at round
\(t+1\).

Consequently,
\[
  \sum_{t=1}^{T}
  \bell^{(t)}(x_p^{(t+1)})
  =
  \sum_{t=1}^{T}\bell^{(t)}(a_t)
  =
  (T\theta,\lambda),
\]
and hence
\[
  \left\|
    \sum_{t=1}^{T}
    \bell^{(t)}(x_p^{(t+1)})
  \right\|_p
  \ge
  T\theta .
\]
On the other hand,
\[
  \left\|
    \sum_{t=1}^{T}\bell^{(t)}(c)
  \right\|_p
  =
  \|\bm L^{(T)}(c)\|_p
  =
  T^{1/p}.
\]
Thus \eqref{eq:p-norm-btl-question} implies
\[
  \alpha
  \ge
  \theta T^{1-1/p}
  =
  \Omega\!\left(T^{1-1/p}\right).
\]
\end{proof}

\subsection{The \texorpdfstring{\(p=1\)}{p=1} Case}
\label{app:p-one}

The obstruction above disappears for \(p=1\).
For nonnegative vector losses, the \(\ell_1\)-norm of the cumulative
vector loss is additive across rounds.

Let
\(
  \bell^{(t)}(x)\in\mathbb R_+^m
\)
and define
\(
  x_1^{(t+1)}
  \in
  \operatorname*{arg\,min}_{x\in\mathcal X}
  \left\|
    \sum_{s=1}^{t}\bell^{(s)}(x)
  \right\|_1 .
\)
By coordinate-wise nonnegativity,
\[
  \left\|
    \sum_{s=1}^{t}\bell^{(s)}(x)
  \right\|_1
  =
  \sum_{s=1}^{t}
  \|\bell^{(s)}(x)\|_1 .
\]
Thus \(x_1^{(t+1)}\) is precisely the leader for the scalar losses
\(
  g^{(s)}(x):=\|\bell^{(s)}(x)\|_1
\).
Applying \Cref{lem:scalar-ftl-btl},
for every fixed \(u\in\mathcal X\),
\[
  \sum_{t=1}^{T}
  \|\bell^{(t)}(x_1^{(t+1)})\|_1
  \le
  \sum_{t=1}^{T}
  \|\bell^{(t)}(u)\|_1 .
\]
Using nonnegativity again,
\[
  \left\|
    \sum_{t=1}^{T}
    \bell^{(t)}(x_1^{(t+1)})
  \right\|_1
  \le
  \left\|
    \sum_{t=1}^{T}
    \bell^{(t)}(u)
  \right\|_1 .
\]
Hence the FTL--BTL lemma holds for \(p=1\) with
\(\alpha=1\).

\subsection{Using an
\texorpdfstring{\(\ell_1\)}{l1}-Leader with an
\texorpdfstring{\(\ell_p\)}{lp}-Objective}
\label{app:l1-leader-lp-eval}

The \(p=1\) case also yields a dimension-dependent bound for other
norms if the decisions are generated by the \(\ell_1\)-leader rather
than the \(\ell_p\)-leader.
For every \(\bm z\in\mathbb R_+^m\) and
\(p\in[1,\infty]\),
\[
  \|\bm z\|_p
  \le
  \|\bm z\|_1
  \le
  m^{1-1/p}\|\bm z\|_p,
\]
with the convention \(m^{1-1/\infty}=m\).
Combining these inequalities with the result of
\Cref{app:p-one}, for every fixed \(u\in\mathcal X\),
\[
  \left\|
    \sum_{t=1}^{T}\bell^{(t)}(x_1^{(t+1)})
  \right\|_p
  \le
  \left\|
    \sum_{t=1}^{T}\bell^{(t)}(x_1^{(t+1)})
  \right\|_1
  \le
  \left\|
    \sum_{t=1}^{T}\bell^{(t)}(u)
  \right\|_1
  \le
  m^{1-1/p}
  \left\|
    \sum_{t=1}^{T}\bell^{(t)}(u)
  \right\|_p .
\]
Thus, the \(\ell_1\)-leader gives a valid
\(m^{1-1/p}\) factor when performance is measured by the
\(\ell_p\)-norm.
This is not, however, an extended FTL--BTL lemma for the
\(\ell_p\)-leader itself.
In particular, for \(p=\infty\), the factor is \(m\), which is too
large to yield the logarithmic dependence sought for \OMMT.

\subsection{A Bound in Terms of Distinct Leaders}
\label{app:distinct-leaders}

Although a universal extended FTL--BTL lemma fails, the number of
distinct leaders can provide a sequence-dependent bound.
Let
\[
  \cD_T
  :=
  \{x_p^{(t+1)}:t\in[T]\},
  \qquad
  K_{\mathrm{dist}}
  :=
  |\cD_T|.
\]

\begin{proposition}
\label{prop:distinct-leader}
Let
\(
  x_p^{(t+1)}
  \in
  \operatorname*{arg\,min}_{x\in\mathcal X}
  \left\|
    \sum_{s=1}^{t}\bell^{(s)}(x)
  \right\|_p .
\)
Then, for every fixed \(u\in\mathcal X\),
\[
  \left\|
    \sum_{t=1}^{T}
    \bell^{(t)}(x_p^{(t+1)})
  \right\|_p
  \le
  K_{\mathrm{dist}}
  \left\|
    \sum_{t=1}^{T}
    \bell^{(t)}(u)
  \right\|_p .
\]
\end{proposition}

\begin{proof}
For each \(a\in\cD_T\), define
\[
  I_a
  :=
  \{t\in[T]:x_p^{(t+1)}=a\},
  \qquad
  b_a:=\max I_a,
  \qquad
  \bm S_a
  :=
  \sum_{t\in I_a}\bell^{(t)}(a).
\]
Since the losses are coordinate-wise nonnegative and
\(I_a\subseteq\{1,\ldots,b_a\}\),
\(
  \bm S_a
  \le
  \sum_{s=1}^{b_a}\bell^{(s)}(a)
\)
coordinate-wise, and therefore
\(
  \|\bm S_a\|_p
  \le
  \left\|
    \sum_{s=1}^{b_a}\bell^{(s)}(a)
  \right\|_p .
\)
Since \(a=x_p^{(b_a+1)}\) is a leader used at round \(b_a+1\),
\[
  \left\|
    \sum_{s=1}^{b_a}\bell^{(s)}(a)
  \right\|_p
  \le
  \left\|
    \sum_{s=1}^{b_a}\bell^{(s)}(u)
  \right\|_p
  \le
  \left\|
    \sum_{s=1}^{T}\bell^{(s)}(u)
  \right\|_p ,
\]
where the last inequality follows from coordinate-wise nonnegativity.

Finally,
\(
  \sum_{t=1}^{T}
  \bell^{(t)}(x_p^{(t+1)})
  =
  \sum_{a\in\cD_T}\bm S_a .
\)
Hence, by the triangle inequality,
\[
  \left\|
    \sum_{t=1}^{T}
    \bell^{(t)}(x_p^{(t+1)})
  \right\|_p
  \le
  \sum_{a\in\cD_T}\|\bm S_a\|_p
  \le
  K_{\mathrm{dist}}
  \left\|
    \sum_{t=1}^{T}\bell^{(t)}(u)
  \right\|_p .
\]
\end{proof}

Here, \(K_{\mathrm{dist}}\) counts distinct leaders rather than leader
switches.
For example, the sequence \(a,b,a,b,a,b\) has five switches but only
two distinct leaders.
For \OMMT, however, \(\mathcal X=\Sn\), and hence the worst-case bound
\(
  K_{\mathrm{dist}}\le\min\{T,n!\}
\)
is still too large to provide a useful adversarial guarantee.

\section{Details of the OLVC Reduction}
\label{app:olvc-reduction}

We provide the detailed analysis of the OLVC reduction used in
\Cref{sec:olvc-reduction}.
Recall that
\[
  \bm{V}^{(t)}
  :=
  \sum_{s=1}^{t}\bm{v}^{(s)}(x^{(s)}),
  \qquad
  \bm{V}^{(0)}:=\bm{0},
\]
and that the surrogate loss is
\[
  f^{(t)}(x)
  =
  \left\langle
    \bm{g}^{(t)},\bm{v}^{(t)}(x)
  \right\rangle,
  \qquad
  \bm{g}^{(t)}
  =
  \nabla\psi_\eta(\bm{V}^{(t-1)}).
\]

The smooth-max function \(\psi_\eta\) defined in
\Cref{sec:olvc-reduction} satisfies the following properties.

\begin{enumerate}
  \item \textbf{Additive approximation.}
  For every \(\bm{u}\in\mathbb{R}_{\ge0}^d\),
  \begin{equation}
  \label{eq:app-additive-approx-inf}
    \|\bm{u}\|_\infty
    \le
    \psi_\eta(\bm{u})
    \le
    \|\bm{u}\|_\infty
    +
    \frac{\ln d}{\eta}.
  \end{equation}

  \item \textbf{Gradient norm bound.}
  For every \(\bm{u}\in\mathbb{R}_{\ge0}^d\),
  \begin{equation}
  \label{eq:app-gradient-norm}
    \|\nabla\psi_\eta(\bm{u})\|_1
    \le 1 .
  \end{equation}

  \item \textbf{Gradient stability.}
  For every
  \(\bm{u}\in\mathbb{R}_{\ge0}^d\) and
  \(\bm{z}\in[0,1]^d\),
  \begin{equation}
  \label{eq:app-gradient-stability}
    \nabla\psi_\eta(\bm{u}+\bm{z})
    \le
    e^\eta\nabla\psi_\eta(\bm{u})
    \qquad
    \text{coordinate-wise}.
  \end{equation}
\end{enumerate}

\begin{myclaim}
\label{clm:app-one-step}
For every round \(t\),
\[
  \psi_\eta(\bm{V}^{(t)})
  -
  \psi_\eta(\bm{V}^{(t-1)})
  \le
  e^\eta f^{(t)}(x^{(t)}).
\]
\end{myclaim}

\begin{proof}
Since
\(
  \bm{V}^{(t)}
  =
  \bm{V}^{(t-1)}
  +
  \bm{v}^{(t)}(x^{(t)})
\),
convexity of \(\psi_\eta\) gives
\[
  \psi_\eta(\bm{V}^{(t)})
  -
  \psi_\eta(\bm{V}^{(t-1)})
  \le
  \left\langle
    \nabla\psi_\eta(\bm{V}^{(t)}),
    \bm{v}^{(t)}(x^{(t)})
  \right\rangle .
\]
Since
\(
  \bm{v}^{(t)}(x^{(t)})\in[0,1]^d
\),
\eqref{eq:app-gradient-stability} implies
\[
  \nabla\psi_\eta(\bm{V}^{(t)})
  \le
  e^\eta
  \nabla\psi_\eta(\bm{V}^{(t-1)})
\]
coordinate-wise.
Therefore,
\[
  \psi_\eta(\bm{V}^{(t)})
  -
  \psi_\eta(\bm{V}^{(t-1)})
  \le
  e^\eta
  \left\langle
    \nabla\psi_\eta(\bm{V}^{(t-1)}),
    \bm{v}^{(t)}(x^{(t)})
  \right\rangle
  =
  e^\eta f^{(t)}(x^{(t)}).
\]
\end{proof}

Define
\[
  \phi_\eta(\bm{u})
  :=
  \sum_{i=1}^{d}e^{\eta u_i},
  \qquad
  \bm{u}\in\mathbb{R}_{\ge0}^d .
\]
Then
\[
  \psi_\eta(\bm{u})
  =
  \frac{1}{\eta}\ln\phi_\eta(\bm{u}),
  \qquad
  \nabla\psi_\eta(\bm{u})
  =
  \frac{1}{
    \eta\phi_\eta(\bm{u})
  }
  \nabla\phi_\eta(\bm{u}).
\]

\begin{lemma}
\label{lem:app-exp-aggregation}
For any
\(\bm{u}\in\mathbb{R}_{\ge0}^d\) and
\(\bm{z}^{(1)},\ldots,\bm{z}^{(K)}
\in\mathbb{R}_{\ge0}^d\),
\[
  \sum_{k=1}^{K}
  \left[
    \phi_\eta\bigl(\bm{u}+\bm{z}^{(k)}\bigr)
    -
    \phi_\eta(\bm{u})
  \right]
  \le
  \phi_\eta
  \left(
    \bm{u}
    +
    \sum_{k=1}^{K}\bm{z}^{(k)}
  \right)
  -
  \phi_\eta(\bm{u}).
\]
\end{lemma}

\begin{proof}
By the definition of \(\phi_\eta\),
\[
  \phi_\eta\bigl(\bm{u}+\bm{z}^{(k)}\bigr)
  -
  \phi_\eta(\bm{u})
  =
  \sum_{i=1}^{d}
  e^{\eta u_i}
  \left(
    e^{\eta z_i^{(k)}}-1
  \right).
\]
Hence
\[
  \sum_{k=1}^{K}
  \left[
    \phi_\eta\bigl(\bm{u}+\bm{z}^{(k)}\bigr)
    -
    \phi_\eta(\bm{u})
  \right]
  =
  \sum_{i=1}^{d}
  e^{\eta u_i}
  \sum_{k=1}^{K}
  \left(
    e^{\eta z_i^{(k)}}-1
  \right).
\]
Since \(z_i^{(k)}\ge0\),
\[
  \sum_{k=1}^{K}
  \left(
    e^{\eta z_i^{(k)}}-1
  \right)
  \le
  \prod_{k=1}^{K}
  e^{\eta z_i^{(k)}}-1
  =
  e^{\eta\sum_{k=1}^{K}z_i^{(k)}}-1 .
\]
Therefore,
\[
  \sum_{k=1}^{K}
  \left[
    \phi_\eta\bigl(\bm{u}+\bm{z}^{(k)}\bigr)
    -
    \phi_\eta(\bm{u})
  \right]
  \le
  \sum_{i=1}^{d}
  e^{\eta u_i}
  \left(
    e^{\eta\sum_{k=1}^{K}z_i^{(k)}}-1
  \right),
\]
which is exactly
\[
  \phi_\eta
  \left(
    \bm{u}
    +
    \sum_{k=1}^{K}\bm{z}^{(k)}
  \right)
  -
  \phi_\eta(\bm{u}).
\]
\end{proof}

\begin{lemma}
\label{lem:app-comparator-infty}
Let
\(
  x^\star
  \in
  \operatorname*{arg\,min}_{x\in\mathcal X}
  \left\|
    \sum_{t=1}^{T}\bm{v}^{(t)}(x)
  \right\|_\infty .
\)
Then
\[
  \sum_{t=1}^{T}f^{(t)}(x^\star)
  \le
  \left(
    e^{\eta\OPT_T^{\mathrm{OLVC}}}-1
  \right)
  \left(
    e^\eta
    \sum_{t=1}^{T}f^{(t)}(x^{(t)})
    +
    \frac{1}{\eta}
  \right).
\]
\end{lemma}

\begin{proof}
By the definition of \(x^\star\),
\begin{equation}
\label{eq:app-comparator-opt}
  \left\|
    \sum_{t=1}^{T}
    \bm{v}^{(t)}(x^\star)
  \right\|_\infty
  =
  \OPT_T^{\mathrm{OLVC}}.
\end{equation}

Since
\[
  \nabla\psi_\eta(\bm{u})
  =
  \frac{1}{
    \eta\phi_\eta(\bm{u})
  }
  \nabla\phi_\eta(\bm{u}),
\]
convexity of \(\phi_\eta\) gives, for every \(t\),
\begin{equation}
\label{eq:app-comparator-single-step}
  f^{(t)}(x^\star)
  \le
  \frac{
    \phi_\eta
    \bigl(
      \bm{V}^{(t-1)}
      +
      \bm{v}^{(t)}(x^\star)
    \bigr)
    -
    \phi_\eta(\bm{V}^{(t-1)})
  }{
    \eta\phi_\eta(\bm{V}^{(t-1)})
  } .
\end{equation}

Define
\[
  A_t
  :=
  \phi_\eta
  \bigl(
    \bm{V}^{(t-1)}
    +
    \bm{v}^{(t)}(x^\star)
  \bigr)
  -
  \phi_\eta(\bm{V}^{(t-1)}).
\]
For \(q=1,\ldots,T\), define
\[
  a_q
  :=
  \begin{cases}
  \displaystyle
  \frac{1}{
    \eta\phi_\eta(\bm{V}^{(q-1)})
  }
  -
  \frac{1}{
    \eta\phi_\eta(\bm{V}^{(q)})
  },
  & q=1,\ldots,T-1,
  \\[2ex]
  \displaystyle
  \frac{1}{
    \eta\phi_\eta(\bm{V}^{(T-1)})
  },
  & q=T.
  \end{cases}
\]
Since
\(
  \bm{V}^{(q)}
  \ge
  \bm{V}^{(q-1)}
\)
coordinate-wise, \(a_q\ge0\).
Moreover,
\begin{equation}
\label{eq:app-comparator-telescoping}
  \sum_{q=t}^{T}a_q
  =
  \frac{1}{
    \eta\phi_\eta(\bm{V}^{(t-1)})
  }.
\end{equation}
Combining
\eqref{eq:app-comparator-single-step} and
\eqref{eq:app-comparator-telescoping} gives
\begin{equation}
\label{eq:app-comparator-rearrange}
  \sum_{t=1}^{T}f^{(t)}(x^\star)
  \le
  \sum_{q=1}^{T}
  a_q
  \sum_{t=1}^{q}A_t .
\end{equation}

We next bound
\(
  \sum_{t=1}^{q}A_t
\).
For every \(t\le q\),
\(
  \bm{V}^{(t-1)}
  \le
  \bm{V}^{(q-1)}
\)
coordinate-wise.
Moreover, the map
\(
  \bm{u}
  \mapsto
  \phi_\eta
  \bigl(
    \bm{u}+\bm{v}^{(t)}(x^\star)
  \bigr)
  -
  \phi_\eta(\bm{u})
\)
is coordinate-wise nondecreasing.
Hence
\[
  A_t
  \le
  \phi_\eta
  \bigl(
    \bm{V}^{(q-1)}
    +
    \bm{v}^{(t)}(x^\star)
  \bigr)
  -
  \phi_\eta(\bm{V}^{(q-1)}).
\]
Applying \Cref{lem:app-exp-aggregation},
\begin{equation}
\label{eq:app-comparator-aggregation}
  \sum_{t=1}^{q}A_t
  \le
  \phi_\eta
  \left(
    \bm{V}^{(q-1)}
    +
    \sum_{t=1}^{q}
    \bm{v}^{(t)}(x^\star)
  \right)
  -
  \phi_\eta(\bm{V}^{(q-1)}).
\end{equation}

By \eqref{eq:app-comparator-opt},
\[
  \sum_{t=1}^{q}
  \bm{v}^{(t)}(x^\star)
  \le
  \sum_{t=1}^{T}
  \bm{v}^{(t)}(x^\star)
  \le
  \OPT_T^{\mathrm{OLVC}}\bm{1}
\]
coordinate-wise.
Therefore,
\begin{equation}
\label{eq:app-comparator-prefix-bound}
  \sum_{t=1}^{q}A_t
  \le
  \left(
    e^{\eta\OPT_T^{\mathrm{OLVC}}}-1
  \right)
  \phi_\eta(\bm{V}^{(q-1)}).
\end{equation}
Substituting this into
\eqref{eq:app-comparator-rearrange},
\begin{equation}
\label{eq:app-comparator-before-last-sum}
  \sum_{t=1}^{T}f^{(t)}(x^\star)
  \le
  \left(
    e^{\eta\OPT_T^{\mathrm{OLVC}}}-1
  \right)
  \sum_{q=1}^{T}
  a_q\phi_\eta(\bm{V}^{(q-1)}).
\end{equation}

By the definition of \(a_q\),
\begin{equation}
\label{eq:app-comparator-last-sum-expand}
  \sum_{q=1}^{T}
  a_q\phi_\eta(\bm{V}^{(q-1)})
  =
  \sum_{q=1}^{T-1}
  \frac{
    \phi_\eta(\bm{V}^{(q)})
    -
    \phi_\eta(\bm{V}^{(q-1)})
  }{
    \eta\phi_\eta(\bm{V}^{(q)})
  }
  +
  \frac{1}{\eta}.
\end{equation}
For \(q=1,\ldots,T-1\), convexity of \(\phi_\eta\) gives
\[
  \phi_\eta(\bm{V}^{(q)})
  -
  \phi_\eta(\bm{V}^{(q-1)})
  \le
  \left\langle
    \nabla\phi_\eta(\bm{V}^{(q)}),
    \bm{v}^{(q)}(x^{(q)})
  \right\rangle .
\]
Thus,
\[
  \frac{
    \phi_\eta(\bm{V}^{(q)})
    -
    \phi_\eta(\bm{V}^{(q-1)})
  }{
    \eta\phi_\eta(\bm{V}^{(q)})
  }
  \le
  \left\langle
    \nabla\psi_\eta(\bm{V}^{(q)}),
    \bm{v}^{(q)}(x^{(q)})
  \right\rangle .
\]
Using \eqref{eq:app-gradient-stability},
\(
  \nabla\psi_\eta(\bm{V}^{(q)})
  \le
  e^\eta
  \nabla\psi_\eta(\bm{V}^{(q-1)})
\)
coordinate-wise, and hence
\[
  \left\langle
    \nabla\psi_\eta(\bm{V}^{(q)}),
    \bm{v}^{(q)}(x^{(q)})
  \right\rangle
  \le
  e^\eta f^{(q)}(x^{(q)}).
\]
Therefore,
\begin{equation}
\label{eq:app-comparator-last-sum-bound}
  \sum_{q=1}^{T}
  a_q\phi_\eta(\bm{V}^{(q-1)})
  \le
  e^\eta
  \sum_{q=1}^{T}f^{(q)}(x^{(q)})
  +
  \frac{1}{\eta}.
\end{equation}

Combining
\eqref{eq:app-comparator-before-last-sum} and
\eqref{eq:app-comparator-last-sum-bound} proves
\[
  \sum_{t=1}^{T}f^{(t)}(x^\star)
  \le
  \left(
    e^{\eta\OPT_T^{\mathrm{OLVC}}}-1
  \right)
  \left(
    e^\eta
    \sum_{t=1}^{T}f^{(t)}(x^{(t)})
    +
    \frac{1}{\eta}
  \right).
\]
\end{proof}

\subsection{Proof of \Cref{thm:olvc-reduction}}
\label{app:proof-olvc-reduction}

\begin{proof}
Let
\(
  S_T
  :=
  \sum_{t=1}^{T}f^{(t)}(x^{(t)}).
\)
By \eqref{eq:surrogate-regret},
\[
  \E[S_T]
  \le
  \E\left[
    \min_{x\in\mathcal X}
    \sum_{t=1}^{T}f^{(t)}(x)
  \right]
  +
  R_T
  \le
  \E\left[
    \sum_{t=1}^{T}f^{(t)}(x^\star)
  \right]
  +
  R_T .
\]
Applying \Cref{lem:app-comparator-infty} and taking expectations gives
\begin{equation}
\label{eq:app-surrogate-recursion}
  \E[S_T]
  \le
  \left(
    e^{\eta\OPT_T^{\mathrm{OLVC}}}-1
  \right)
  \left(
    e^\eta\E[S_T]
    +
    \frac{1}{\eta}
  \right)
  +
  R_T .
\end{equation}

By the choice of \(\eta\) in
\Cref{thm:olvc-reduction},
\(
  e^{\eta\OPT_T^{\mathrm{OLVC}}}-1
  \le
  \gamma
\)
and
\(
  e^\eta
  \le
  \frac{c}{\gamma}.
\)
Hence
\(
  \left(
    e^{\eta\OPT_T^{\mathrm{OLVC}}}-1
  \right)e^\eta
  \le c .
\)
Using these inequalities in
\eqref{eq:app-surrogate-recursion} gives
\begin{equation}
\label{eq:app-surrogate-bound}
  \E[S_T]
  \le
  \frac{\gamma}{(1-c)\eta}
  +
  \frac{1}{1-c}R_T .
\end{equation}

By \Cref{clm:app-one-step},
\[
  \psi_\eta(\bm{V}^{(T)})
  -
  \psi_\eta(\bm{0})
  \le
  e^\eta
  \sum_{t=1}^{T}f^{(t)}(x^{(t)}).
\]
Since
\(
  \psi_\eta(\bm{0})=(\ln d)/\eta
\)
and
\(
  \ALG_T^{\mathrm{OLVC}}
  =
  \|\bm{V}^{(T)}\|_\infty
  \le
  \psi_\eta(\bm{V}^{(T)})
\),
we obtain
\begin{equation}
\label{eq:app-olvc-potential-bound}
  \ALG_T^{\mathrm{OLVC}}
  \le
  \frac{\ln d}{\eta}
  +
  e^\eta
  \sum_{t=1}^{T}f^{(t)}(x^{(t)}).
\end{equation}
Taking expectations and applying
\eqref{eq:app-surrogate-bound},
\[
  \E\!\left[
    \ALG_T^{\mathrm{OLVC}}
  \right]
  \le
  \frac{\ln d}{\eta}
  +
  e^\eta
  \left(
    \frac{\gamma}{(1-c)\eta}
    +
    \frac{1}{1-c}R_T
  \right).
\]
Since \(e^\eta\le c/\gamma\),
\begin{equation}
\label{eq:app-olvc-before-eta-substitution}
  \E\!\left[
    \ALG_T^{\mathrm{OLVC}}
  \right]
  \le
  \left(
    \ln d+\frac{c}{1-c}
  \right)
  \frac{1}{\eta}
  +
  \frac{c}{\gamma(1-c)}R_T .
\end{equation}

Finally, the definition of \(\eta\) implies
\begin{equation}
\label{eq:app-eta-inverse-bound}
  \frac{1}{\eta}
  \le
  \frac{
    \OPT_T^{\mathrm{OLVC}}
  }{
    \ln(1+\gamma)
  }
  +
  \frac{1}{
    \ln(c/\gamma)
  } .
\end{equation}
Indeed, if
\(
  \OPT_T^{\mathrm{OLVC}}>0
\),
then
\(
  \frac{1}{\eta}
  =
  \max
  \left\{
    \frac{1}{\ln(c/\gamma)},
    \frac{\OPT_T^{\mathrm{OLVC}}}{\ln(1+\gamma)}
  \right\},
\)
and the claimed bound follows from
\(
  \max\{a,b\}\le a+b
\).
If
\(
  \OPT_T^{\mathrm{OLVC}}=0
\),
then
\(
  \eta=\ln(c/\gamma)
\),
so \eqref{eq:app-eta-inverse-bound} holds with equality.
Substituting
\eqref{eq:app-eta-inverse-bound} into
\eqref{eq:app-olvc-before-eta-substitution} yields
\[
  \E\!\left[
    \ALG_T^{\mathrm{OLVC}}
  \right]
  \le
  \frac{
    \ln d+c/(1-c)
  }{
    \ln(1+\gamma)
  }
  \OPT_T^{\mathrm{OLVC}}
  +
  \frac{
    \ln d+c/(1-c)
  }{
    \ln(c/\gamma)
  }
  +
  \frac{c}{
    \gamma(1-c)
  }
  R_T .
\]
This proves \Cref{thm:olvc-reduction}.
\end{proof}

\fi

\end{document}